\documentclass[preprint,11pt,authoryear]{elsarticle}
\usepackage{lipsum}
\makeatletter
\def\ps@pprintTitle{%
 \let\@oddhead\@empty
 \let\@evenhead\@empty
 \def\@oddfoot{\centerline{\thepage}}%
 \let\@evenfoot\@oddfoot}
\makeatother

\usepackage{xcolor}
\usepackage{subfig}
\usepackage[ruled,vlined]{algorithm2e}

\usepackage[T1]{fontenc}
\usepackage{tikz}
\usetikzlibrary{arrows,shapes,decorations,automata,backgrounds,petri}
\usepackage{setspace}
\usepackage[a4paper, total={6in, 8in}]{geometry}

\usepackage{amssymb}
\usepackage{url}
\usepackage{amsthm}
\usepackage{comment}

\usepackage{amsmath}
  {
     \newtheorem{assumption}{Assumption}
     
     \newtheorem{theorem}{Theorem}

     \newtheorem{proposition}{Proposition}
  }

\usepackage{lineno}

\usepackage{setspace}

\begin{document}

\begin{frontmatter}



\title{Optimal discharge of patients from intensive care via a data-driven policy learning framework}


\author[1,2]{Fernando Lejarza} 
\ead{flejarza@dascena.com} 
\author[1]{Jacob Calvert\corref{cor1}}
\ead{jake@dascena.com} 
\author[1]{Misty M Attwood} 
\ead{mattwood@dascena.com} 
\author[1]{Daniel Evans} 
\ead{devans@dascena.com} 
\author[1]{Qingqing Mao} 
\ead{qingqing@dascena.com} 
\cortext[cor1]{Corresponding author}
\address[1]{Dascena, Inc., 12333 Sowden Road, Suite B, Houston, TX 77080-2059, USA\fnref{label1}}
\address[2]{McKetta Department of Chemical Engineering, The University of Texas at Austin, Austin, TX 78712, USA\fnref{label2}}

\begin{abstract}
Clinical decision support tools rooted in machine learning and optimization can provide significant value to healthcare providers, including through better management of intensive care units. In particular, it is important that the patient discharge task addresses the nuanced trade-off between decreasing a patient's length of stay (and associated hospitalization costs) and the risk of readmission or even death following the discharge decision. This work introduces an end-to-end general framework for capturing this trade-off to recommend optimal discharge timing decisions given a patient's electronic health records. A data-driven approach is used to derive a parsimonious, discrete state space representation that captures a patient's physiological condition. Based on this model and a given cost function, an infinite-horizon discounted Markov decision process is formulated and solved numerically to compute an optimal discharge policy, whose value is assessed using off-policy evaluation strategies. Extensive numerical experiments are performed to validate the proposed framework using real-life intensive care unit patient data. 
\end{abstract}



\begin{keyword}
Healthcare \sep Intensive care unit \sep Markov Decision Process \sep Policy learning \sep Machine learning 

\end{keyword}

\end{frontmatter}

\def\ps@pprintTitle{%
 \let\@oddhead\@empty
 \let\@evenhead\@empty
 \def\@oddfoot{}%
 \let\@evenfoot\@oddfoot}
\makeatother

\section{Introduction}
\label{s:intro}

The demand for accurate recommendation systems for discharging patients poses a substantial challenge in intensive care units (ICU). Advanced patient needs, coupled with the greater demand for staff and resources as well as limited capacity, elevate the costs of healthcare delivery in ICUs to levels much greater than those in generic medical-surgical hospital wards \citep{badawi2012readmissions, halpern2010critical}. These factors intuitively motivate the discharge of patients from ICUs as soon as their health status no longer demands such extensive resources \citep{gordon2018prevalence}. Nonetheless, patients who are readmitted after being prematurely discharged from ICUs, or whose health status declines rapidly after discharge, impose an even greater burden on hospitals and have disproportionately high mortality rates \citep{badawi2012readmissions,mcneill2020impact}. A recent review of hospital mortality studies reported that readmitted patients can have readmission mortality rates as high as 40 percent compared to non-readmitted mortality rates of 3.6 to 8.4 percent \citep{mcneill2020impact}.

Furthermore, hospitals in the United States have an additional financial motivation to reduce rates of premature or improper discharge, as they can be penalized through the Hospital Readmissions Reduction Program (HRRP) if they fail to meet federal standards \citep{wasfy2017readmission}. HRRP penalizes the majority of American hospitals for substandard readmission rates, up to a maximum penalty rate of one percent of the hospital’s Medicare base payments \citep{hrrp2013}. Ensuring efficient patient discharge with minimal probability of ICU readmission is therefore an important priority for critical care providers, which incentivizes the creation and deployment of decision support tools that can help clinicians optimize the timing of patient discharge \citep{levin2021machine}. This priority has become even more pressing with the ongoing COVID-19 public health crisis, which has imposed a dramatically higher burden on critical care worldwide \citep{arabi2021covid}. The development of well-designed machine learning and optimization schemes could meet this need by interpreting ICU patients’ electronic health records (EHR) data in real-time, assessing their probability of readmission, and providing clinicians with concise assessments to assist them with these difficult, high-stakes decisions. Algorithms and decision support tools may also encourage clinicians to discharge patients outside of normal rounding hours, allowing for greater fluidity in patient turnover \citep{mcwilliams2019towards}.

The fields of operations research (OR) and artificial intelligence (AI) are well informed of the challenges in developing algorithms that meet the complex demands of this clinical problem. ICU patient populations are highly heterogeneous with respect to disease state and comorbidity \citep{forte2021identifying}, and the relatively low proportions of ICU patients who are readmission cases may yield imbalance in data sets \citep{loreto2020early}. The death of patients after discharge from ICUs, either at home or at other healthcare facilities outside the purview of data collection, is a competing risk in algorithm development that may often be driven by the same factors that would indicate the need for readmission \citep{loreto2020early}. The increased mortality rate within ICUs relative to other hospital wards also generates a higher attrition rate for data collection \citep{wilcox2019challenges, mcwilliams2019towards}. Despite these setbacks, tools for classifying patients by their probability of readmission continue to be developed with steadily improving performance  \citep{mcwilliams2019towards, mcneill2020impact, balshi2020modified, czajka2020validation, loreto2020early, levin2021machine}.

In this work we present an unsupervised machine learning approach for representing patients' physiological conditions, inherently reflecting mortality and readmission risks. This health state representation is then used to formulate an infinite horizon Markov decision process in order to numerically compute an optimal discharge policy for a given cost function. The proposed policy is validated using recently developed off-policy evaluation algorithms on a large scale real-life ICU patient data set. We discuss the resulting policy's implication to hospital management, as well as its interpretability relative to the clinicians' decisions.

\section{Literature overview}
\label{s:lit_overview}

 Several machine learning methods to predict unplanned hospital or ICU ward readmission within  48 hours \citep{desautels2017prediction}, and 7, 14, or 30 days  \citep{maali2018predicting, jaotombo2020machine, lo2021predictive} or within the same hospitalization \citep{rojas2018predicting} have been published. A recent review of the application of machine learning in predicting hospital readmission identified 43 relevant studies employing a variety of modeling methods \citep{huang2021application}. Additional studies have focused on predicting multiple outcomes simultaneously including both post-discharge mortality and readmission \citep{badawi2012readmissions, campbell2008predicting, ouanes2012model,thoral2021explainable}, as well as length of stay and readmission \citep{hilton2020personalized}.

Machine learning models leveraging EHR data suffer from several limitations including: potential exclusion of patients due to missing data, varying feature consistency and availability across electronic medical record systems, and the temporal availability of data codes for classifying patients \citet{temple2015predicting}. Because of these reasons models developed to predict patient discharge or time to discharge have been studied to a lesser extent. \citet{mcwilliams2019towards} proposed random forest and logistic regression classifiers to identify discharge-ready patients based on a variety of demographic and EHR features. The developed models significantly outperformed a much more conservative nurse-led discharge criteria \citep{knight2003nurseleddischarge} used as benchmark, although the relatively high rate of false-positives indicated the necessity of further development before clinical deployment. \citet{temple2015predicting} developed a random forest model to identify neonatal ICU patients with high discharge likelihood within a time window of two to ten days.  A random forest regression model was developed by \citet{cuadrado2019pursuing} to predict the discharge time, i.e., the length of stay or time to discharge, of ICU patients based on 49 clinically relevant variables and values from 15 different assessment scores. Several artificial neural network methods have been used to develop prediction algorithms for length of stay of ICU patients \citep{Gholipour2015annlos}, and specifically for cardiac patients \citep{Rowan2007anncardiac, lafaro2016nncardiac}. \citet{safavi2019development} developed a feedforward neural network model to predict daily inpatient surgical care discharges within a 24 hour time window. Further, the framework proposed by \citet{safavi2019development} was used to quantitatively identify clinical milestones to recovery and barriers to discharge that, respectively, indicate progression or postponement towards discharge. 

As noted in several studies \citep{rubenfeld2014many, stelfox2015scoping, mcwilliams2019towards}, identifying patients that are suitable for ICU discharge is a complex task due to the myriad of factors that can drive such recommendations (e.g., patient demands, hospital management culture, procedures favored by a given clinician, resource availability, profit motivations, etc.). In discharge prediction models, determining the risk level or  threshold above which a patient can be safely discharged is challenging and involves balancing true-positives and false-positive rates, depending on the needs of the clinical team. For example, clinicians may be interested in a lower threshold to identify all potential discharge patients including those that may not be ready yet for discharge, or hospital bed managers that may prefer higher specificity so they can more accurately balance beds for incoming patient demand \citep{safavi2019development}. Limitations in the aforementioned prediction algorithms, such as restrictive feature sets and high false positive rates \citep{mcwilliams2019towards}, as well as missing hospital-level factors such as ICU bed availability and hospital census data \citep{rojas2018predicting} have fostered the development of \textit{prescriptive} (i.e., recommending a specific course of action as opposed to predicting a likely outcome) artificial intelligence frameworks. 

The focus in this work is to develop patient-specific data-driven \textit{discharge recommendation policies}, rather than to train models to predict either patient time to discharge, or readmission and out-of-hospital mortality instances. The former approach optimizes over custom objective functions considering patient health dynamics and reflecting clinician and hospital management preferences to recommend a specific course of action, and provide more easily translatable and clinically usable health-based decision rules. Several related modelling schemes have been proposed for constructing optimal policies for sequential decision-making to address various types of medical problems \citep{schaefer2005modeling, bennett2013artificial, komorowski2018artificial}. These studies introduce frameworks typically based on different variants of Markov decision process (MDP) models  \citep{bertsekas2012dynamic, puterman2014markov} that can effectively capture the dynamic nature of patients' health condition and provide optimal clinical decisions within the limits of the structural assumptions invoked in defining the system states, actions, transitions, and costs. These often called ``artificial intelligence clinicians'' (AIC) \citep{komorowski2018artificial} simulate clinical decision-making through consideration of multiple dynamic factors, including patient health conditions and demographics, as well as hospital system-related characteristics and costs. For example,  \citet{bennett2013artificial} developed a general purpose (non-disease specific) computational environment to sequentially simulate multiple decision paths for replicating and even improving clinician decision-making in real-time. The framework was evaluated in a chronic care setting, in which the ``clinical utility'' (capturing the cost-effectiveness of treatments reflected by outcomes and costs) was optimized improving patient outcomes by 30-35\% at a much lower cost of care. A plethora of frameworks have been proposed for constructing treatment policies for different specific chronic medical conditions including HIV \citep{shechter2008optimal}, diabetes \citep{denton2009optimizing}, anemia \citep{gaweda2005individualization}, and breast cancer \citep{ayvaci2012effect}, among several others.  Recently, \citet{komorowski2018artificial} proposed an AIC to compute real-time optimal sepsis treatment strategies consisting of intravenous fluids and vasopressor dosages to maximize patients' 90-day expected survival.  \citet{komorowski2018artificial} demonstrated that the AIC consistently outperformed the clinician policy even on external validation data corresponding to different hospitals, and confirmed that the AIC decisions were also highly interpretable. Comprehensive reviews of successful applications of MDPs and reinforcement learning for medical treatment policies can be found in \cite{schaefer2005modeling, steimle2017markov, yu2019reinforcement}.

Similar frameworks related to optimal patient discharging have been less studied in the literature. \citet{kreke2008modeling} proposed a finite-time MDP framework that modelled when to discharge patients with pneumonia-related sepsis to maximize their expected survival rate. Through structural and computational analyses, \citet{kreke2008modeling} showed that for specific problem instances the optimal discharge strategy follows a non-stationary control-limit-type policy, implying that the level of illness at which it is optimal to discharge a patient changes over the course of the hospital stay. The Sequential Organ Failure Assessment (SOFA) Score is used to represent the patient health states, which is recognised to be a simplistic metric that may fail to capture the complex set of clinical features available to represent the patient's physiological condition. \citet{chan2012optimizeicu} developed a demand-driven index-based greedy policy that discharges an ICU patient with the lowest ``criticality index'' when a new patient arriving in the ICU must be accommodated. The proposed state-space reflected the total number of ICU patients at a given time period, where each patients was labeled from a finite set of  ailments/health conditions. The resulting greedy policy balanced patient health-related costs (such as physiological deterioration leading to readmission or post-discharge mortality) with system related costs (relating to increased length of stay and capacity-limited ICU resources), and was theoretically proofed and empirically demonstrated to be near-optimal in settings of low ICU utilization. Numerical experiments demonstrated that the resulting policy reduced the patient readmission load by nearly 30\% respective to selected benchmark policies. 
\citet{ouyang2020allocation} considered a related ICU bed allocation problem during periods of high patient demand. The objective was to minimize the long-run average expected mortality rate, and the model considered two possible health states: critical and highly critical condition. Seven different policies were evaluated, and numerical simulations showed that the proposed framework consistently reduced patient mortality risk. \citet{shi2021timing} developed a large-scale MDP to optimize ICU patient flow management, patient quality of care, and patient outcome (readmission risk). The proposed model featured a personalized readmission prediction model to dynamically determine which patients to discharge on each day based on current ICU occupancy. Under the assumption that costs related to ICU congestion and discharge (dependent on the expected number of readmissions for a given a number of discharges) have quadratic structure, \citet{shi2021timing} developed an efficient linear decision rule approximation for the optimal policy which was empirically demonstrated to decrease readmission risk from 32\% to 28\% by increasing the average length of stay from 3.33 days to 3.55 days. 

To our knowledge, our work is the first to provide an end-to-end framework that includes modeling patient health states from high dimensional time series EHR data, computing optimal discharge policies based on different cost trade-offs, and evaluating their performance and ICU management implications in real ICU patient health trajectories. The main contributions of the present work are: 
\begin{enumerate}
    \item A flexible, data-driven prescriptive framework for making discharge decisions such that a given cost function (reflecting the trade-off between hospitalization expenses, readmission penalties, and out-of-hospital mortality rate) is minimized. 
    \item A clustering-based approach for identifying discrete patient health states from high-dimensional continuous EHR data, that inherently reflects the relationship between physiological conditions and outcomes such as mortality and readmission.  
    \item Numerical validation experiments performed using a large ICU database (MIMIC-III, \cite{johnson2016mimic}) collected from over 400,000 adult patients in the United States. 
    \item Off-policy evaluation algorithms and results to derive statistically significant performance guarantees of the proposed discharge policy relative to selected policy benchmarks. 
    \item The resulting policy can effectively lower hospital readmissions $\sim$2\%, while at the same time reducing the average length of stay by  $\sim$1 day relative to the clinician policy in a hold-out testing set. 
\end{enumerate}

\section{Materials and methods}
\label{s:problem_statement}

\subsection{Data set description and processing} 
\label{ss:dataset}

We developed and validated the proposed framework using the Medical Information Mart for Intensive Care (MIMIC-III, version 1.3) data set \citep{johnson2016mimic}. MIMIC-III consists of EHR data from the ICU visits of over 400,000 patients, collected from a large, tertiary care hospital in Boston, Massachusetts, from 2001 to 2012. Each encounter included vital signs, medication information, laboratory measurements, and diagnostic codes, among other data. Because the MIMIC-III data are de-identified, this study qualified as a non-human subject study according to the definition of human subjects research put forth in 45 CFR 46, and was exempt from Institutional Review Board approval. 

Encounters were subjected to the inclusion procedure depicted in Figure S1. Encounters consisted of measurements of clinical variables, including demographic information, vital signs, and laboratory tests (Tables S1 and S2). The first step of the inclusion procedure removed encounters with no measurement of more than half of the vital signs and laboratory measurements.

Measurements were binned into 12-hour periods. Specifically, the measurements of each clinical variable within each 12-hour period were aggregated either by averaging the measurements (for most vital signs and laboratory tests) or by adding them (e.g., for urine output, drug and fluid administration). Missing measurements were imputed by carrying forward the last observation. If no aggregated measurement was available in a period, then the latest available observation of a patient's health state (prior to that time period) was applied. 
If values were missing for medications or treatments, it was assumed that none were administered or performed and they were imputed with zeros. Outliers were replaced by capping the aggregated feature measurements to their associated 99.9 and 0.1 percentiles. 
If any missing values remained (patients not having an entire measurement available, or the first observation of a given measurement time series) iterative imputation was performed based on Bayesian Ridge regression as the default estimator \citep{pedregosa2011scikit}.

\subsection{Markov decision process framework}
\label{ss:MDP_structure}

We considered the problem of whether or not to discharge a patient from the ICU based on the information presently available regarding their condition (Table S2). The objective was to minimize a given cost function comprising two components:
\begin{itemize}
    \item the daily cost of hospitalization and implied treatments; and
    \item the $T$-day expected rate of unsuccessful discharges, where typical values of $T$ include 30, 60, and 90 days \citep{kreke2008modeling}.
\end{itemize} We considered the discharge of a patient to be {\em unsuccessful} if the patient was readmitted to the ICU within $T$ days of discharge, or if the patient died within $T$ days of discharge.

The cost function was intended to capture the trade-off between keeping a patient in intensive care to improve the odds of a successful discharge at the expense of higher operating costs to the hospital, higher medical bills to the patient, and higher ICU bed occupancy. The latter may be undesirable, especially in the context of the COVID-19 pandemic. The relative weights of each component of the cost function are to be determined on a case-by-case basis, in such a way that the resulting discharge policy reflects the hospital's ICU management strategy. We formalized the problem using the notation of \citet{bertsekas2012dynamic}.

\subsubsection{Time representation}

According to the model, discharge decisions were made every 12 hours. The discharge task was formulated as an infinite-horizon discounted MDP, that captured the transitions of patients through discrete health ``states'' which represented the condition of the patient in consecutive 12-hour time periods indexed by $t \in \{1,2,\dots\}$. While only the $T$-day time window following discharge was considered, a patient could in principle be kept in the ICU indefinitely if their health condition was not seen to improve or if a high weight was placed on the readmission component of the cost function. This property will be elaborated in Section \ref{s:numerical_sim}. 

\subsubsection{State and action space}

The MDP state space, $\mathcal{S}$, was the set of health states  $\{1,2,\dots,H,SD,UD\}$, where states $1, 2, \dots, H$ represented the patient's physiological condition during their ICU stay, and where the (absorbing) states $SD$ and $UD$ corresponded to successful and unsuccessful outcomes following the decision to discharge a patient. Denote the condition of a given patient in the $t$\textsuperscript{th} 12-hour time period of their stay by $x_t \in \mathcal{S}$. In Section \ref{s:health_states}, we detail how states $\{1,2,\dots,H\}$ were learned from data, and defined in terms of the information in Table S2. We remark that, while continuous state-space representations can retain more information about a patient \citep{raghu2017continuous}, their use can lead to computational tractability issues, which motivates the use of discrete representations \citep{kreke2008modeling, komorowski2018artificial}. 

The action space $\mathcal{A}$ consisted of $K$, the action to keep a patient in the ICU for at least one more time period and $D$, the action to discharge them from it. In fact, the action $u_t$ taken at the end of time period $t$ depended on the health state $x_t$. If $x_t \in \{1,2,\dots,H\}$, then $u_t \in \mathcal{A}$ but, if $x_t \in \{SD,UD\}$, $u_t = \emptyset$ (no further action could be taken).

\begin{assumption}
This framework does not consider any actions/decisions with respect to medical treatments, laboratory measurements or other procedures received by the patients during the ICU stay. When estimating transition probabilities $P(\tilde{x}|x,K)$ from data, an implicit policy is learned regarding the procedures used to treat the patient's condition when the health state is $x$. This treatment policy is tacitly followed every time that the proposed discharge policy recommends keeping the patient in care.
\end{assumption}

\subsubsection{Transition probabilities}

Let $P(\tilde{x}|x,u)$ denote the probability that a certain patient's health state is $\tilde{x}$ in the next time period, given that the current state is $x$ and action $u$ is taken. These probabilities were estimated using data corresponding to health transition observed for a large number of different patients during their stay in the ICU. By definition, if $\tilde x \in \{SD,UD\}$, then $P(\tilde x | x, u) = 1$ if and only if $x = \tilde x$ and $u = \emptyset$. Note also that a patient could not transition to $SD$ or $UD$ unless discharged, i.e. $P(SD|x,K) = P(UD|x,K) = 0$.

\subsubsection{Costs}

We formulated the MDP problem in terms of cost minimization. Denote the immediate cost of taking action $u$ in state $x$ by $g(x, u)$. If $x \in \{1,2,\dots,H\}$, then $g(x, K)$ was the cost of care for one time period. The cost $g(UD,\emptyset)$, or simply $g(UD)$, corresponded to a penalty for an unsuccessful discharge. A reward in the form of $g(SD) < 0$ reflected a favorable discharge outcome. As the choice of cost function $g$ will vary across different hospitals and according to ICU bed utilization preferences, the numerical results presented below explored the effects of different cost functions on the behavior of the optimal policy.

\subsubsection{Infinite horizon discounted MDP} 

Together, the states, actions, transition probabilities, and costs define an MDP. The objective of solving an MDP is to find a policy, or a mapping from a patient's health state to a discharge decision for all decision epochs that minimizes a measure of long-term, expected, discounted costs for a patient's stay in the ICU. While discounting in the context of this problem had no practical significance, time-discounted MDPs have been studied extensively and benefit from favorable convergence properties and multiple solution algorithms \citep{puterman2014markov}.

For an initial state $x_0$ and discount factor $\alpha \in (0,1)$, we are interested in finding the policy $\pi = \{\mu_0, \mu_1, \dots \}$ in terms of $\mu_t: \mathcal{S} \rightarrow \mathcal{A}$, that minimized the cost
\begin{equation}
\label{eq:discounted_extected_costs}
J_{\pi}(x_0) = \lim_{N\rightarrow \infty} \mathbb{E} \left [ \sum_{t=0}^{N-1} \alpha^{k} g(x_t, \mu_t(x_t))  \right ].
\end{equation}
The optimal policy, for some initial health state $x_0$, is the one with the least cost $J^*(x_0)$ given by
\begin{equation}
\label{eq:optimal_costs}
J^{*}(x_0) = \min_{\pi\in \Pi}  J_\pi(x_0) \;\; \forall x_0 \in \mathcal{S},
\end{equation}
where $\Pi$ is the set of all admissible policies. A policy is said to be stationary if it has the form of $\pi=\{\mu,\mu,\dots\}$, in which case $\pi$ is simply referred to as $\mu$. The reader is referred to \citet{bertsekas2012dynamic} and \citet{puterman2014markov} for a comprehensive discussion on MDPs.

\section{Modeling patient's health states and dynamics} 
\label{s:health_states}

\subsection{Clustering electronic health records data}

We assigned to each 12-hour time period (preceding discharge) of each patient's stay a health state $\{1,2,\dots,H\}$, on the basis of the information in Table S2 available during that period. This involved (i) determining a number $H$ of health states to define and then (ii) defining each health state in terms of the clinical variables. Concerning (i), if too few of health states are defined, then they will reflect the patient's condition poorly \citep{alagoz2010markov}. However, if too many health states are defined, then the subsequent policy optimization step will not be computationally tractable, and there will not be enough data to accurately estimate the transition probabilities between them. Concerning (ii), unsupervised machine learning---specifically, clustering algorithms---have been used to identify groups of patients with similar medical characteristics \citep{forte2021identifying,el2009length,alashwal2019application,komorowski2018artificial}.

We employed the $k$-means clustering algorithm to identify health states, where $k$ corresponds to the number of possible states, $H$. In brief, the algorithm seeks to partition the collection of data from each 12-hour time period in the training data into $k$ clusters, such that each example belongs to the cluster with the nearest centroid, which were found by minimizing the intra-cluster distance. Formally, the objective was to solve 
\begin{equation}
\label{eq:k_means}
\min_{\textbf{Y}} \sum_{i=1}^k \sum_{\textbf{y}\in Y_i} \| \textbf{y} - \textbf{c}_i \|^2,
\end{equation}
where $\| \cdot \|$ denotes the Euclidean distance and $\textbf{y}$ denotes the examples in the training set. The constituent sets of $\textbf{Y} = \{Y_1, \dots, Y_k \}$ are the clusters, and $\textbf{c}_i$ is the centroid of the points of $Y_i$. Solving \eqref{eq:k_means} directly is NP-hard, and thus heuristic algorithms, like Lloyd's algorithm, are used instead to iteratively find the local optimum of \eqref{eq:k_means} and corresponding cluster centroids.  

\begin{figure}[htb]
\centering
\subfloat[]{\label{fig:clusters_400_IHD}\includegraphics[width=0.5\linewidth]{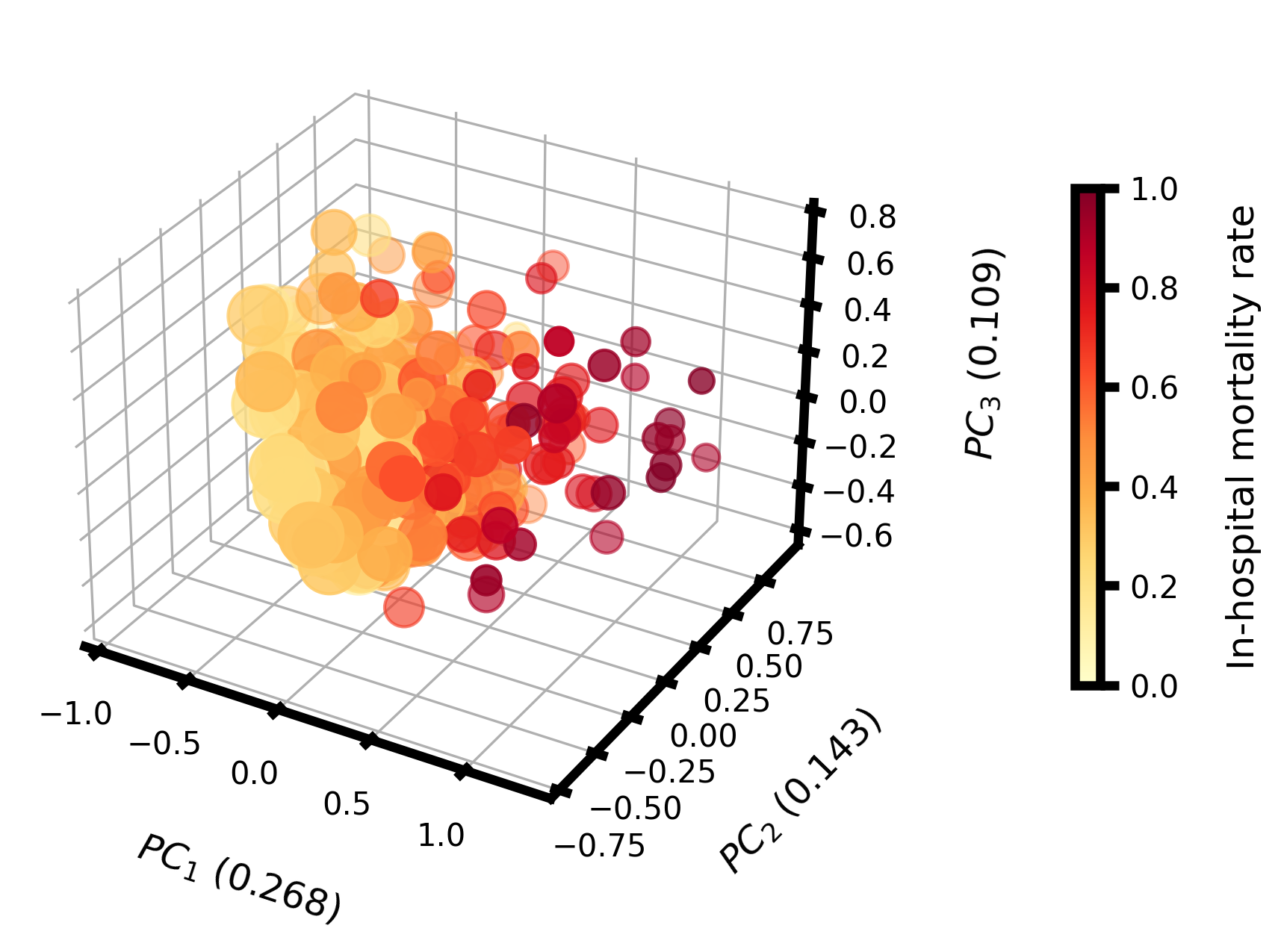}} \hfill
\subfloat[]{\label{fig:clusters_400_RA}\includegraphics[width=.5\linewidth]{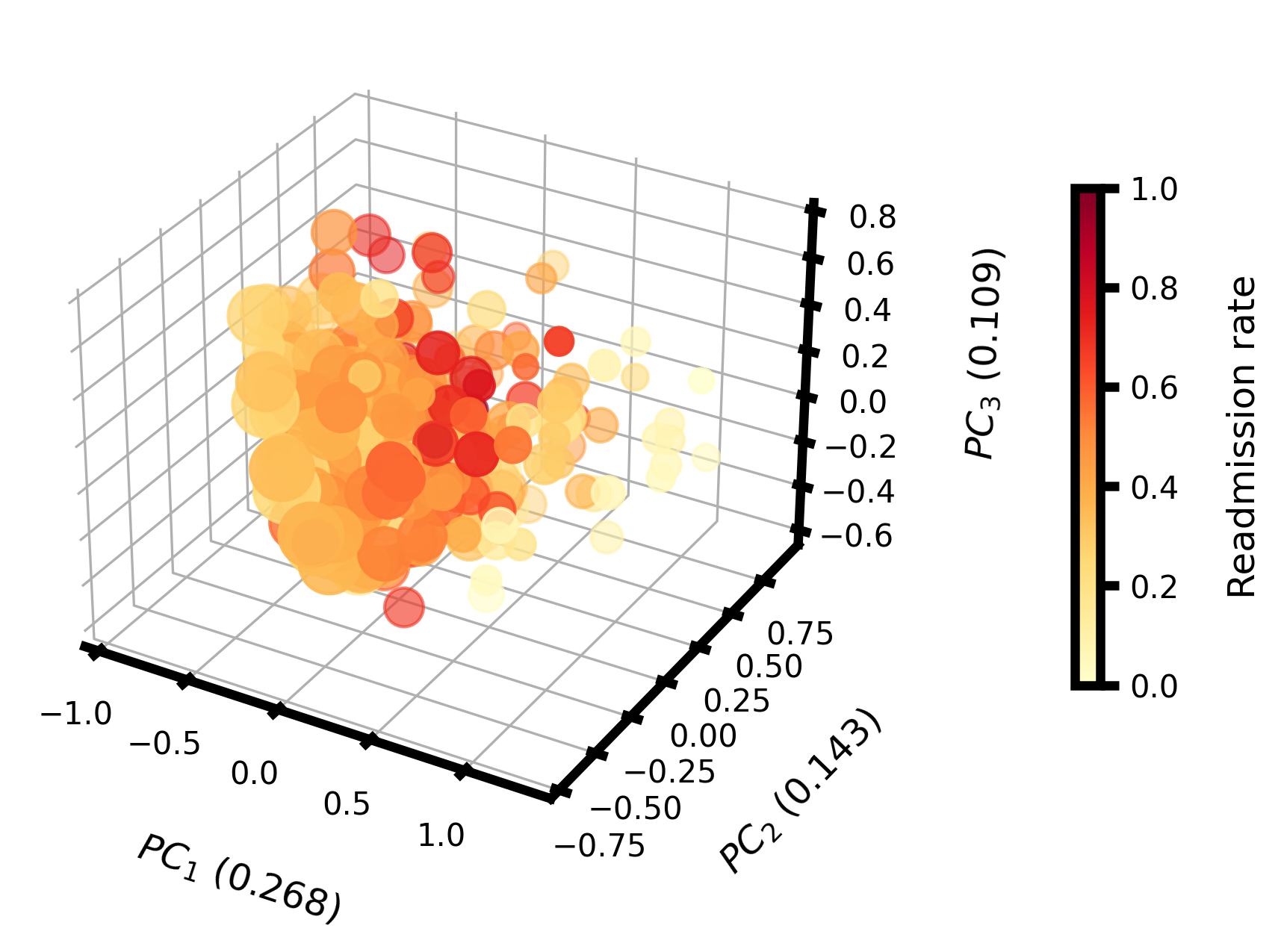}}\par
\caption{Health states identified via $k$-means clustering using $k=400$ and shown for the three principal components with the highest explained variance shown in parentheses in the corresponding axis labels. Marker color indicates the (A) average mortality rate and (B) average 30-day readmission rate for the corresponding cluster. Marker size corresponds to the number of examples assigned to the cluster.} 
\label{fig:clustering_results}
\end{figure}

The training data were scaled using the min-max scaler prior to training the clustering algorithm. The number of possible health states $H$, equivalent to the number of clusters $k$, was chosen to be 400 according to the elbow method \citep{syakur2018integration} (Figure S2). This number of clusters was large enough to ensure separability between patients based on their distinct clinical features, but small enough such that there was a sufficient number of transitions observed in the training set in order to estimate the transition probabilities empirically. The identified clusters are shown in Figure \ref{fig:clustering_results}, where the axes represent the three principal components with the highest explained variance. The marker size in the scatter plots reflects the number of examples in the entire training data set assigned to each cluster, and the color scale reflects the average mortality and average 30-day readmission of the assigned examples to each cluster.

\begin{figure}[htb]
\centering\includegraphics[scale=0.5]{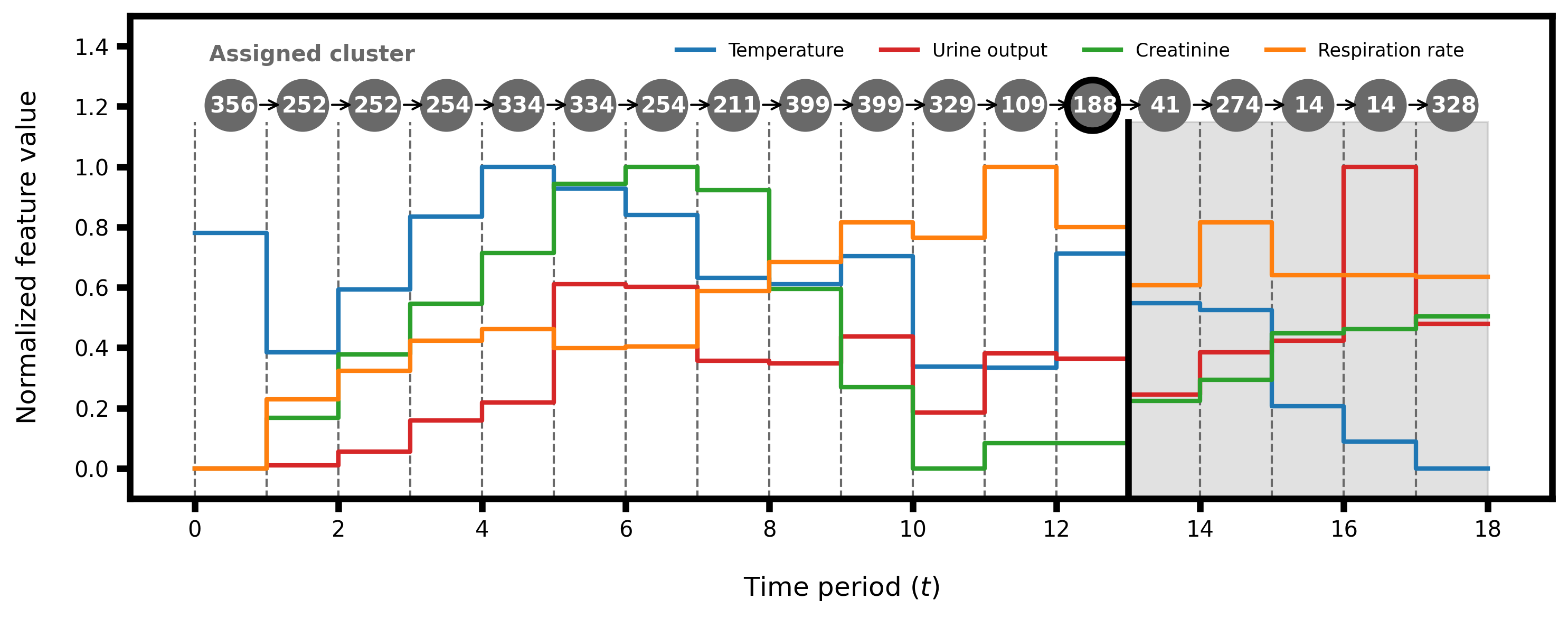}
\caption{Example time series of EHR data and associated health states (for simplicity only four  clinical variables are shown).}
\label{fig:EHR_cluster}
\end{figure}

We observed a gradient of average in-hospital mortality rate for the health states visualized in Figure \ref{fig:clusters_400_IHD}, despite the fact that mortality information was not provided to the clustering algorithm, which is unsupervised by definition. This observation supports the validity of the cluster assignment as a reflection of the severity of the patient's condition. 
We observed an analogous gradient in the average 30-day readmission rate of the learned health states (Figure \ref{fig:clusters_400_RA}). The high mortality health states corresponded the states with the lowest 30-day readmission rates because the patients assigned to these states either died in the hospital, and therefore never had the chance to be readmitted, or were in such critical condition that their condition must have improved before being discharged. Figure \ref{fig:clusters_400_RA} indicates that the highest readmission rates typically resulted from patients that were in moderate mortality clusters, perhaps after their condition improved but before they had fully recuperated, which may have resulted in readmission. An example time series 12-hour periods of EHR data and associated health states are shown in Figure \ref{fig:EHR_cluster}. The bold, vertical line indicates the time period at which the optimal policy recommended discharge (details on how this policy is computed are presented shortly), and the grey-shaded region corresponds to the continuation of the stay under the clinician policy. 

\subsection{Empirical estimation and validation of transition probabilities}
\label{ss:TPs}

After determining the 400 health states, we estimated the transition probabilities between health states. For example, the transition probability $P(\tilde{x}|x,K)$ was estimated by counting all observations in the training set in which a transition from state $x$ to state $\tilde{x}$ occurred, and dividing the count by the total number of observations of state $x$ for which the patient was not discharged. To hedge against discharging patients with very small readmission rates but high mortality rates, $P(UD|x,D)$ was estimated by also taking into consideration instances of in-hospital mortality. That is, $P(UD|x,D)$ was estimated by counting three events: (1) the number of 30-day readmissions when patients were discharged with health state $x$, (2) the number of $30$-day out-of-hospital mortality instances when patients were discharged with health state $x$, and (3) the number of in-hospital deaths for patients with last observed health state $x$. The sum of these three observations was divided by the total number of discharges, plus the number of  in-hospital deaths observed for patients with health state $x$. Similarly, $P(SD|x,D)$ was estimated as $1-P(UD|x,D)$, as $UD$ and $SD$ are the only two possible outcomes when a patient is discharged. 

\begin{figure}[htb]
\centering\includegraphics[scale=0.45]{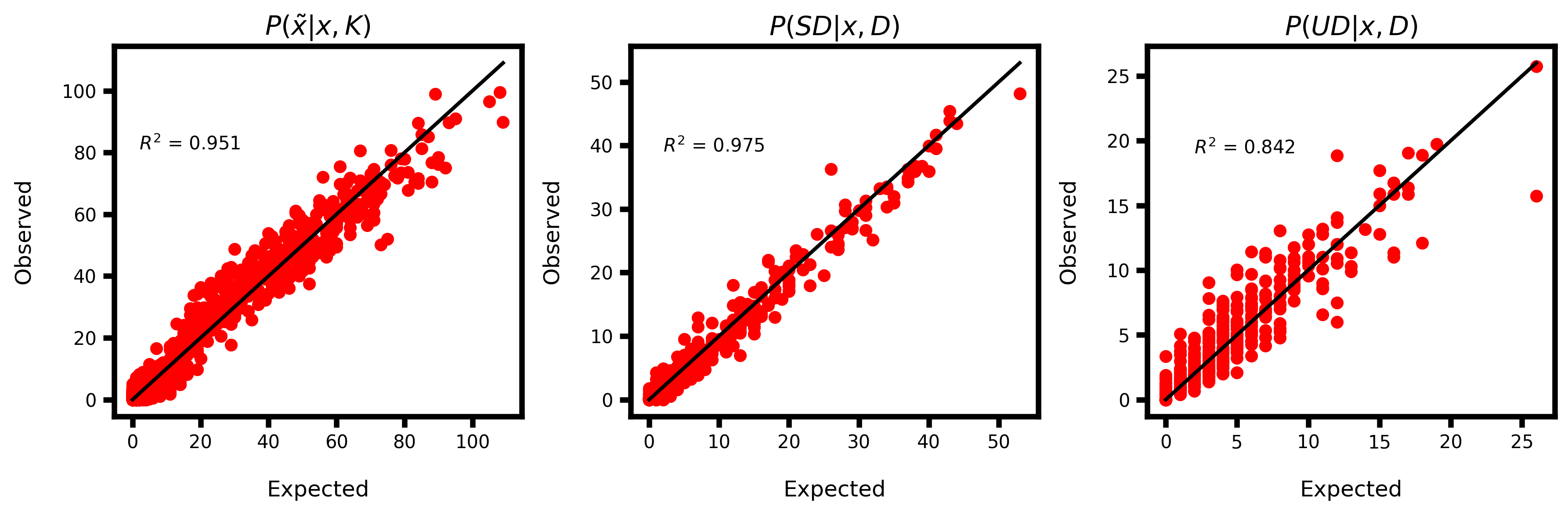}
\caption{Validation results for transition probabilities estimated from training data and evaluated on a hold-out test set.}
\label{fig:TP_validation}
\end{figure}

We evaluated the quality of the estimated transition probability matrices using the observations in the hold-out testing data set. Given the number of transitions contained in the test set, we used the estimates of $P(\tilde{x}|x,K)$, $P(SD|x,D)$ and $P(UD|x,D)$ to compute the expected number of observations of all possible transitions (e.g., from state $x$ to state $\tilde{x}$ if the patient was kept, or from $x$ to $SD$ or $UD$ if they were discharged). We compared this expected number against the number of transitions observed in the testing set, to assess the goodness of fit of the empirical transition matrices. The validation results for all three transition probability matrices are shown in Figure \ref{fig:TP_validation}. The estimated transition probabilities generalized well to the testing set as indicated by the high coefficient of determination computed for each transition matrix. Table S3 lists the median $R^2$ and interquartile range obtained from repeating the clustering and validation process with 50 random train-test splits of the entire data set. 

We further validated the estimated transition probability matrix $P(\tilde{x}|x,K)$ by simulating the first exit time from each state, as in \citet{komorowski2018artificial}. For each initial health state $x$, we conducted 500 random simulations of the dynamics, to observe the length of time spent at health state $x$ before transitioning to a different health state $\tilde{x}$. The results, shown in Figure S3, confirmed that this time is roughly exponentially distributed for all states. An exponential fit of the times for each state, resulted in a median $R^2$ coefficient of 0.988 and an interquartile range of (0.996-0.999). 

\section{Discharge policy computation and performance evaluation} 
\label{s:policy_computation}

\subsection{Policy characterization}

In the preceding sections, we defined the MDP components $\{\mathcal{S},\mathcal{A},P,g,\alpha\}$, estimated the parameters from the training data, and validated the estimates using a hold-out test set. We turn our focus to identifying the policy which minimizes \eqref{eq:discounted_extected_costs}. First, we state the  assumptions and properties that are critical for the numerical algorithms we used to solve the MDP. The reader is referred to \citep{puterman2014markov,bertsekas2012dynamic} for a comprehensive analysis of these properties, as well as the ensuing computational aspects.

\begin{assumption}[{\citep[Assumption $D$]{bertsekas2012dynamic}}] 
\label{ass:boundedness}
The costs per stage $g(x,u)$ are bounded, that is $|g(u,x)| < M$ for all $(x,u) \in \mathcal{S} \times \mathcal{A}$ where $M$ is some scalar, and the discounting factor satisfies $0 < \alpha < 1$.
\end{assumption}

Assumption \ref{ass:boundedness} is used to derive necessary and sufficient conditions for the optimality of a policy and the convergence of numerical methods \citep{bertsekas2012dynamic}. It holds if the rewards and penalties for successful and unsuccessful discharges respectively are finite, because $\mathcal{S}$ and $\mathcal{A}$ are finite sets, and because the costs associated with hospitalization are bounded. The assumption that $\alpha \in (0,1)$ ensures that the series in \eqref{eq:discounted_extected_costs} converges.

\begin{proposition}[{\citep[Proposition 1.2.2]{bertsekas2012dynamic}}]
\label{prop:bellman}
The optimal cost function $J^*$ satisfies Bellman's optimality equations given by $J^*=\mathcal{T}J^*$, where the mapping $\mathcal{T}$ is given by 
\begin{equation}
\label{eq:mapping}
    (\mathcal{T}J)(x) = \min_{u\in \mathcal{U}(x)} \mathbb{E} \left [ g(x,u) + \alpha J(\tilde{x}) \right ]
\end{equation}
Furthermore, $J^*$ is unique within the class of all bounded functions. 
\end{proposition}
Proposition \ref{prop:bellman} establishes that the optimal cost function is the fixed point of mapping $\mathcal{T}$ defined in \eqref{eq:mapping}. Bellman's optimality equations for health state $x \in \{1,\dots,H\}$ is given by 
\begin{equation}
\label{eq:bellman_discharge}
\begin{aligned}
    J(x) = \min \bigg \{  &  g(x,K) +  \alpha \sum_{\tilde{x} \in \{1,2,\dots,H\} } P(\tilde{x}|x,K) J(\tilde{x}), \\ 
    & g(x,D) +  \alpha \sum_{\tilde{x} \in \{SD,UD \} } P(\tilde{x}|x,D)  J(\tilde{x})  \bigg \}.
\end{aligned}
\end{equation}
Bellman's equations associated to the post-discharge absorbing states $x\in\{SD,UD\}$ are given by 
\begin{equation}
\label{eq:bellman_discharge_2}
\begin{aligned}
    J(x) = \frac{g(x)}{1-\alpha}.
\end{aligned}
\end{equation}

For every stationary policy $\mu$ the associated cost function satisfies  $J_\mu = \mathcal{T}_\mu J_\mu$, where the mapping $\mathcal{T}_\mu$ is given by 
\begin{equation}
    (\mathcal{T}_\mu J)(x) = \mathbb{E} \left [ g(x,\mu(x)) + \alpha J(\tilde{x}) \right ]
\end{equation} 
 Based on these results, we provide the conditions for which the discharge policy's optimality and existence are guaranteed, which is critical for understanding the numerical solution schemes discussed subsequently in Section \ref{ss:policy_comp}. 

\begin{proposition}[{\citep[Proposition 1.2.3]{bertsekas2012dynamic}}]
\label{prop:optimality}
A stationary policy $\mu$ is optimal if and only if $\mu(x)$ attains the minimum in Bellman's equations $J^*=\mathcal{T}J^*$ for each $x \in \mathcal{S}$, i.e., $\mathcal{T}J^*=\mathcal{T}_\mu J^*$
\end{proposition}

\begin{theorem}[{\citep[Theorem 6.2.10]{puterman2014markov}}]
\label{th:existance}
Assume $\mathcal{S}$ is discrete and $\mathcal{A}$ is finite for each  $x\in\mathcal{S}$, then there exists an optimal deterministic stationary policy that solves the optimality equation $J^*=\mathcal{T}J^*$. 
\end{theorem}

Theorem \ref{th:existance} establishes the existence of an optimal discharge policy, while Proposition \ref{prop:optimality} gives a way to identify it in terms of Bellman's equations. We proceed to discuss the algorithm we used to compute the optimal policy. 

\subsection{Policy computation}
\label{ss:policy_comp}

We used policy iteration \citep{bertsekas2012dynamic} to find a numerical solution to the discounted cost problem \eqref{eq:discounted_extected_costs}. We chose it because the finiteness of the state and action spaces, and the satisfaction of Assumption \ref{ass:boundedness} guaranteed that policy iteration would terminate in a finite number of steps \citep{bertsekas2012dynamic}. Because the number of states was not particularly large, we did not expect computational complexity to be an obstacle, and thus preferred it over other approaches, like value iteration. The premise of the policy iteration algorithm is to sequentially compute stationary policies using Bellman's equations, such that the cost of each policy is lower relative to the preceding one. The steps involved in policy iteration are outlined in Algorithm \ref{alg:policy_iter}.

\begin{algorithm}[h]
\SetAlgoLined
 \textbf{Inputs:} MDP elements \{$\mathcal{S},\mathcal{A},P,g,\alpha$\}, initial policy guess $\mu^0$ \\
 \For{$k\in\{0,1,2,\dots\}$}{
 Policy evaluation: compute $J_{\mu^k}$ by solving $(I-\alpha P_{\mu^k})J_{\mu^k}=g_{\mu^k}$ \\ 
 Policy improvement: compute $\mu^{k+1}$ such that $\mathcal{\mathcal{T}}_{\mu^{k+1}}J_{\mu^k}=\mathcal{T}J_{\mu^k}$ \\ 
\If{$J_{\mu^k} = \mathcal{T} J_{\mu^k}$}{
 \Return $\mu^k$
}
}
 \caption{Policy iteration \citep{bertsekas2012dynamic}}
  \label{alg:policy_iter}
\end{algorithm}

The policy evaluation step involves solving a system of linear equations to compute the value function $J_{\mu^k}$ for the current best available policy $\mu^k$. $P_{\mu^k}$ and $g^{\mu^k}$ denote, respectively, the transition probabilities and the cost per stage evaluated for the current policy, and $I$ denotes the identity matrix of appropriate dimensions. We note that policy evaluation is susceptible to the curse of dimensionality, as the problem size increases with the number of state variables, and other numerical algorithms and effective approximations (e.g., \citep{de2003linear}) might be preferable for MDPs with larger state spaces. Once the current value function has been computed, policy improvement is performed by applying the dynamic programming mapping $\mathcal{T}$ defined in \eqref{eq:mapping} on the value function $J$ to compute the subsequent, lower-cost policy. The algorithm terminates once the value function no longer improves, which implies convergence to a fixed point and to the optimal policy (Proposition \ref{prop:optimality}).

\subsection{Policy cost estimation}
\label{ss:policy_eval}

The policy learning framework presented in this work uses a clinician policy ($\mu_{CP}$) implicitly represented in the training and testing data to respectively learn and estimate the value of a different policy ($\mu_{OP}$), which is optimal with respect to the given cost function. Because the execution of a bad policy can be costly and even dangerous in practical settings, particularly pertaining to ICUs in which erroneous discharge decisions can be fatal, it is important to establish the performance of $\mu_{OP}$ off-line using an available testing set of trajectories $\mathcal{D} = \{\mathcal{H}_1,\mathcal{H}_2,\dots, \mathcal{H}_n\}$ where a trajectory $\mathcal{H}$ of length $L$ is a state-action history $\{(x_0,u_0),(x_1,u_1),\dots,(x_{L-1},u_{L-1})\}$ generated by following the clinician's policy $\mathcal{H}\sim\mu_{CP}$.  Approaches for off-policy evaluation  are an active topic of research in the context of related reinforcement learning frameworks \citep{hanna2017bootstrapping, thomas2015high, festor2021enabling}. It should be noted that the  factors driving $\mu_{CP}$ may differ from the cost function chosen to derive $\mu_{OP}$, which is why effective calibration is important to objectively compare the two policies.



We employed the model-based bootstrapping algorithm proposed in \citep[Algorithm 1]{hanna2017bootstrapping} to estimate the value \eqref{eq:discounted_extected_costs} of $\mu_{OP}$ using trajectories in the hold-out test set that follow $\mu_{CP}$. Specifically, given a policy $\mu_{OP_i}$ for $i\in \{1,2,\dots,50\}$ (each $i$ corresponds to a random train-test split of the data and resulting clustering solution), the cost of the policy is estimated by following the test trajectories $\mathcal{D}$ reflecting $\mu_{CP}$, until a deviation between the policies is observed: either $\mu_{OP}$ discharges the patient earlier than $\mu_{CP}$, or $\mu_{CP}$ discharged the patient but $\mu_{OP}$ keeps the patient in care. In these instances, model-based off-policy evaluation  \citep{hanna2017bootstrapping} is performed by simulating future trajectories following $\mu_{OP_i}$ and the empirical transition probabilities estimated using $\mathcal{D}_i$---not the transition probabilities estimated using the training data---to estimate the expected policy costs. If the empirical transition probabilities generalize well to unseen data sets, then this approach reduces the variance of the estimate, at the expense of introducing bias resulting from e.g. data sparsity. The average cost over all trajectories in $\mathcal{D}_i$ was used to estimate the value of $\mu_{OP_i}$. Note that the costs associated with $\mu_{CP}$ are estimated directly by following the trajectories in $\mathcal{D}$. 

To obtain a distribution for the cost associated with a policy $\mu_{OP}$, we performed bootstrapping in combination with the aforementioned off-policy model-based estimation approach. The 95\% confidence upper bound (95\% UB) of the cost of $\mu_{OP_i}$ was determined as proposed in \citep{hanna2017bootstrapping}, by using  bootstrapping with replacement to generate $B$ trajectory sets $\{\tilde{\mathcal{D}}^j_i,\,  j \in \{1,2,\dots,B\}\}$ where $\tilde{\mathcal{D}}_i^j =\{ \mathcal{H}_1^j, \mathcal{H}_2^j, \dots,\mathcal{H}_n^j\}$ such that  $\mathcal{H}_k^j \sim \mathcal{U}(\mathcal{D}_i)$  and where $\mathcal{U}(\cdot)$ is the uniform distribution. Similarly, the 95\% confidence lower bound (95\% LB) corresponding to the implicit clinician policy $\mu_{CP}$ was determined using the same bootstrapped samples. The upper and lower bounds for each policy served to ensure that the worst-case realization of $\mu_{OP}$ still performed better than the best-case realization $\mu_{CP}$ for a number of trained models. The UB and LB were obtained by computing the 97.5 and 2.5 percentiles (denoted by $\mathcal{Q}_{97.5}$ and $\mathcal{Q}_{2.5}$) of the bootstrapped policy cost estimates. This off-policy evaluation procedure was previously demonstrated for clinical decision support tools in the context of sepsis treatment \citep{komorowski2018artificial}. Algorithm \ref{alg:policy_eval} presents this evaluation procedure in detail, and numerical results pertaining to off-policy evaluation are discussed below.

\begin{algorithm}[h]
\SetAlgoLined
 \textbf{Inputs:}  Evaluation policy $\mu_{OP_i}$, set of test trajectories $\mathcal{D}_i$, required number of bootstrap estimates $B$, confidence level $\delta \in [0,1]$ \\
 Estimate transition matrices $\hat{P}_i$ from $\mathcal{D}_i$ \\ 
 \For{$j\in\{1,2,\dots,B\}$}{
    $\tilde{\mathcal{D}}_i^j \leftarrow \{\mathcal{H}_1^j,\mathcal{H}_2^j,\dots, \mathcal{H}_n^j\}$ where $\mathcal{H}_k^j \sim \mathcal{U}(\mathcal{D}_i)$\\
\For{$l\in[1,n]$}{
 Compute $g(\mathcal{H}_l^j)$ for $\mu_{OP_i}$ by simulating model $\{\mathcal{S}, \mathcal{A}, \hat{P}_i, g, \alpha \}$
}
$\hat{J}^j = \frac{1}{n} \sum_{l=1}^n g(\mathcal{H}_l^j) $
}
 \textbf{return} $\mathcal{Q}_{\delta \times 100}(\{\hat{J}^j | j\in\{1,2,\dots,B\} \})$  
 \caption{Model-based off-policy evaluation performed for all policies $\mu_{OP_i}$ for all  $i\in\{1,2,\dots,50\}$ \citep{hanna2017bootstrapping} }
  \label{alg:policy_eval}
\end{algorithm}

\subsection{Random discharge policies for benchmarking}

In addition to comparing $\mu_{OP}$ and $\mu_{CP}$ against one another other, we evaluated their performances against policies that make discharge decisions at random. The first random policy was 
\begin{equation}
    \mu_{RP_1}(x) = \mathcal{U}(\{K,D\}) \;\; \forall \;\; x \in \{1,2,\dots,H\}.
\end{equation}
In words, under $\mu_{RP_1}$, a patient was discharged or kept in care uniformly at random by flipping a fair coin at each time period. This random policy was used to compare the costs relative to $\mu_{OP}$ and $\mu_{CP}$, and was expected to result in higher expected discounted costs. 

Since the proposed policy $\mu_{OP}$ was intended to reduce the number of ICU readmissions and out-of-hospital mortality instances (unsuccessful discharges) by keeping and treating patients in the ICU longer, we wanted to assess how effectively $\mu_{OP}$ selected which patients to discharge and which ones to keep in care. To this end, we considered a second, pseudo-random policy denoted by $\mu_{RP_2}$ that operated as follows: given an encounter trajectory $\mathcal{H}$ that implicitly follows $\mu_{CP}$, $\mu_{RP_2}$ determines whether or not to extend the patient's length of stay randomly with probability $\gamma$ (i.e., by flipping a biased coin). To compare against $\mu_{OP_i}$, $\gamma$ was set to the fraction of patients kept in care by $\mu_{OP_i}$ relative to all the patients observed in the test set $\mathcal{D}_i$. Therefore, by definition, for a given test set of trajectories $\mathcal{D}$, $\mu_{RP_2}$ discharged the same number of patients as $\mu_{OP_i}$, but without considering their health state when making the discharge decision. A summary of the four policies considered in this work are displayed in Table \ref{tb:policies_summary}. 

\begin{table}[htb]
\caption{Summary of all policies considered for numerical simulations and benchmarking.}
\centering
\label{tb:policies_summary} 
\begin{tabular}{c l}
\hline
$\pmb{\mu}$ & \textbf{Policy description} \\  
\hline 
$\mu_{OP}$ & Optimal policy computed via Algorithm \ref{alg:policy_iter} for a given cost function \\ 
$\mu_{CP}$ & Clinician policy implicit in the training and testing data \\ 
$\mu_{RP_1}$ & Purely random policy for arbitrary discharges at any time \\ 
$\mu_{RP_2}$ & Pseudo-random policy with same net discharges as $\mu_{OP}$\\ 
\hline 
\end{tabular} 
\end{table}

\section{Numerical simulations and results}
\label{s:numerical_sim}

To assess the effectiveness of the proposed data-driven policy, the computed discharge policies were evaluated against the ones corresponding to the clinician policy observed in the hold-out test set, as well as to the previously described random policies. The test set trajectories exclusively consisted of encounters in which the clinician was observed to discharge a patient, excluding in-hospital death instances. For the subsequent numerical studies, 50 different clustering solutions were computed using different random train-test splits of the entire data set. We compare the policies on the basis of their resulting expected discounted costs in Section \ref{ss:cost_eval}, and evaluate their performances in terms of the ICU management outcomes of interest (e.g., fraction of unsuccessful discharges and average patient length of stay) in Section \ref{ss:implications_eval}. We discuss the interpretability of the proposed policy in relation to the clinician policy in Section \ref{ss:interpretability}.

\subsection{Off-policy cost evaluation}
\label{ss:cost_eval}

\begin{figure}[htb]
\centering
\subfloat[]{\label{fig:offpolicy_evaluation_conf}\includegraphics[width=0.5\linewidth]{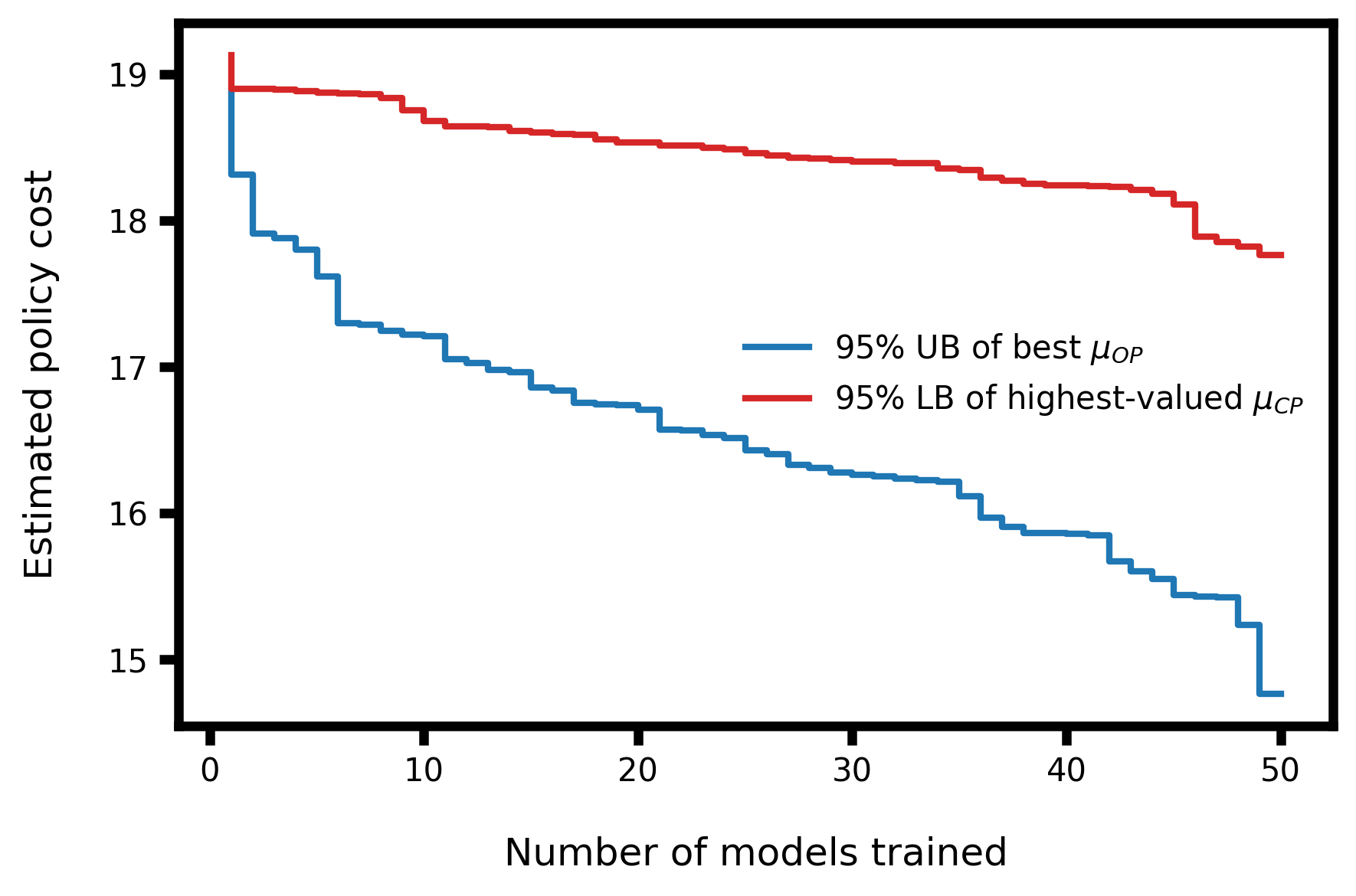}} \hfill
\subfloat[]{\label{fig:offpolicy_evaluation_dist}\includegraphics[width=.5\linewidth]{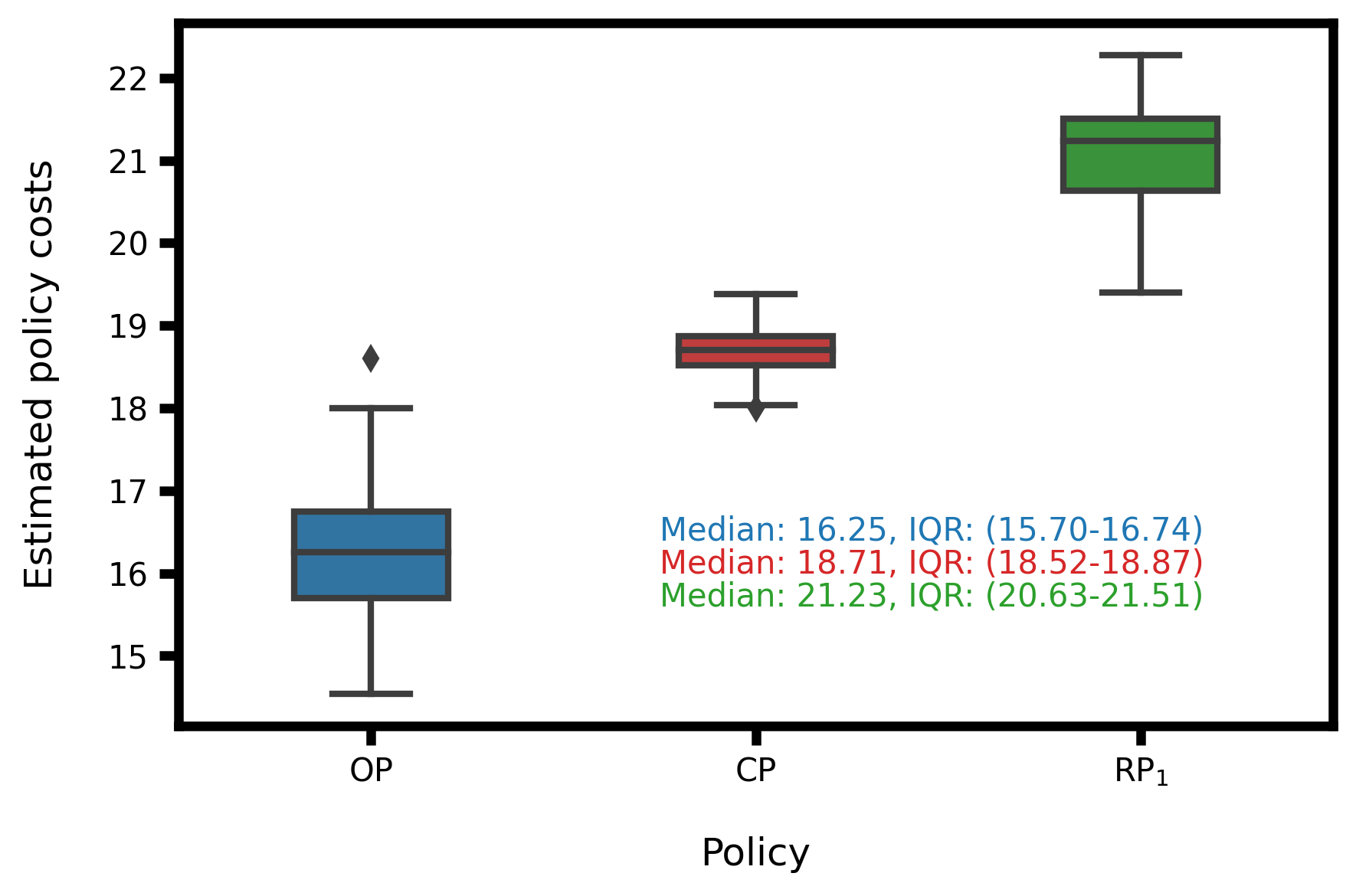}}\par
\caption{(A) Evolution of the 95\% upper confidence bound on the cost of $\mu_{OP}$, and 95\% lower confidence bound on the cost of $\mu_{CP}$ as a function of number of models trained. (B) Distribution of off-policy estimated policy costs. For each training set, $\mu_{OP}$ was  generated using cost function with $g(UD)=3$. Model-based policy evaluation was performed using 100 Monte Carlo simulations, and 2000 bootstrapped repetitions. } 
\label{fig:offpolicy_evaluation}
\end{figure}

For a fixed cost function that is well calibrated---meaning that it results in high costs for patients with high 30-day out-of-hospital mortality and readmission rates, as well as in low costs for patients that have a high rate of successful discharge (Figure S4)---it is important to establish statistical significance on the performance difference between $\mu_{OP}$ and $\mu_{CP}$ on the test set trajectories $\mathcal{D}$. The results corresponding to the model-based off-policy evaluation procedure discussed in Section \ref{ss:policy_eval} are displayed in Figures \ref{fig:offpolicy_evaluation_conf} and \ref{fig:offpolicy_evaluation_dist}. The policy cost estimates for $\mu_{OP}$ for each individual trajectory were computed using 100 random simulations by sampling the transition matrices, and 2000 bootstrapped samples were used to estimate the associated cost distributions (and associated confidence bounds) of $\mu_{OP}$ and $\mu_{CP}$. Figure \ref{fig:offpolicy_evaluation_conf} shows that the 95\% UB on the costs of $\mu_{OP}$ is consistently and largely inferior than the 95\% LB on the costs of $\mu_{CP}$. The difference in the performance between the two policies grew with the number of models trained. The distribution of estimated costs for $\mu_{OP}$ and $\mu_{CP}$ is shown in Figure \ref{fig:offpolicy_evaluation_dist}. 

Furthermore, we used off-policy evaluation to demonstrate that the performance of learned policies is, to some extent, robust to the the number of discrete health states or clusters used to represent a patient's physiological condition. The off-policy evaluation results corresponding to policies computed using 350, 400, and 450 clusters are shown in Table S4.

\subsection{Implications on unsuccessful discharges and length of stay}
\label{ss:implications_eval}

 The policies  associated  with each of the 50 clustering solutions were computed via policy iteration for a range of different cost functions, which were determined by varying the penalty associated with unsuccessful discharges $g(UD)$ while keeping all else constant (that is, $g(x,K)=1$, $g(x,D)=0$, $g(SD)=0$, $\alpha=0.95$). Naturally, the proposed policy $\mu_{OP}$ discharged at most as many ICU stays as were observed in each of the hold-out test sets. Nonetheless, depending on the cost function of choice, $\mu_{OP}$ may recommend that a fraction of the patients are kept in care until their health states are seen to improve (i.e., the patient transitions to a state having a lower probability of unsuccessful discharge), thus leading to a lower total number of discharges than the observations in the test set. To confirm that the improvements stemming from $\mu_{OP}$ were not a mere result of reducing the number of patient discharged, the pseudo-random policy ($\mu_{RP_2}$) was used as a benchmark. 

\begin{figure}[htb]
\centering\includegraphics[scale=0.45]{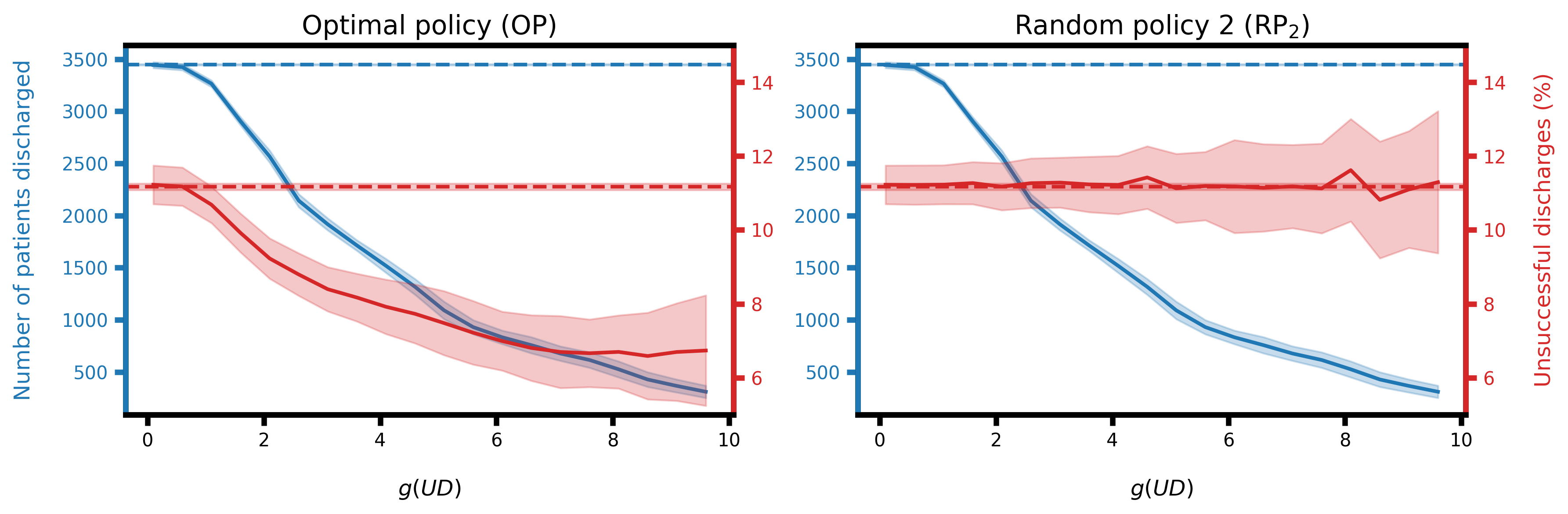}
\caption{Performance curves showing total number of patients discharged and resulting fraction of unsuccessful discharges as a function of cost parameter $g(UD)$ for $\mu_{OP}$ and $\mu_{RP_2}$. Dashed lines indicate the mean values of $\mu_{CP}$, solid lines for $\mu_{OP}$ and RP$_2$ indicate the mean value of 50 trained models evaluated on their corresponding hold-out test sets, and shaded areas correspond to two standard deviations from the respective means.}
\label{fig:perf_curves_OP_RP_CP}
\end{figure}

Results comparing $\mu_{OP}$ and $\mu_{RP_2}$ against $\mu_{CP}$ for a range of $g(UD)$ values are shown in Figure \ref{fig:perf_curves_OP_RP_CP}. In the extreme case when $g(UD)$ approaches zero, meaning there is no penalty associated with unsuccessful discharges, $\mu_{OP}$  discharged all patients to minimize the costs associated with hospitalization. Conversely, increasing $g(UD)$ resulted in a greater fraction of patients kept in care for $\mu_{OP}$ (and by definition also for $\mu_{RP_2}$) relative to $\mu_{CP}$, thus at the expense of a longer length of stay.  Since the discharged encounters are sampled at random from the test set for $\mu_{RP_2}$, the resulting percentage of patients with $UD$ aligned on average with the clinician policy (the average and standard deviation of $p$-values for all instances of $g(UD)$ were 0.67 and 0.28 respectively). On the other hand, for the optimal policy, the fraction of unsuccessful discharges consistently decreased relative to $\mu_{CP}$ and $\mu_{RP_2}$, indicating that the proposed framework  effectively identified patients at a high risk of being unsuccessfully discharged. The higher variance observed in the fraction of $UD$ corresponding to large values of $g(UD)$ for both $\mu_{OP}$ and $\mu_{RP_2}$ was explained by the small sample of discharged patients used to compute the corresponding mean.

\begin{figure}[htb]
\centering\includegraphics[scale=0.45]{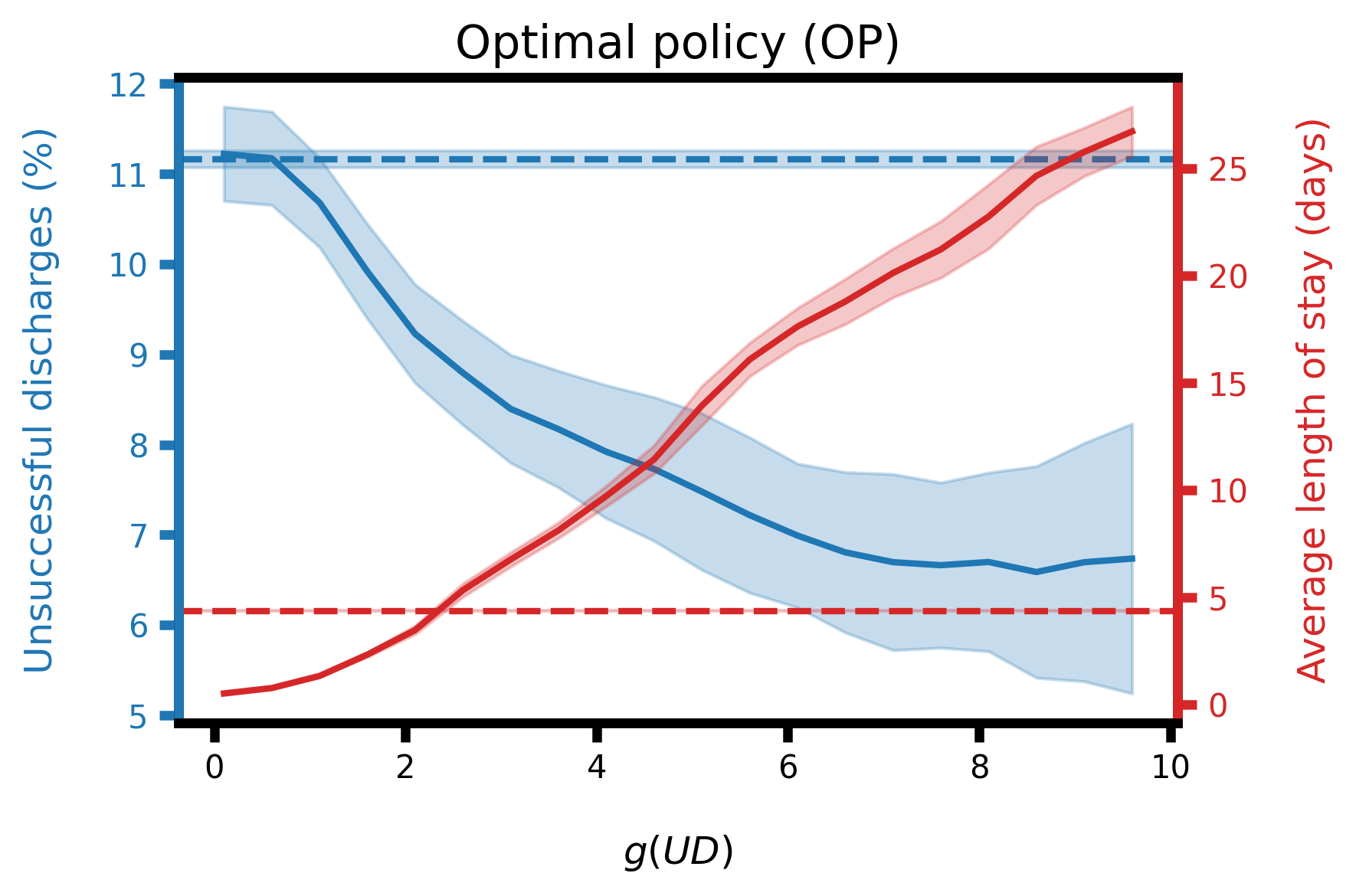}
\caption{Performance curves showing fraction of unsuccessful discharges and average (expected) length of stay evaluated as a function of cost parameter $g(UD)$ for $\mu_{OP}$. Dashed lines indicate the mean values of $\mu_{CP}$, solid lines for $\mu_{OP}$ indicate the mean value of 50 trained models evaluated on their corresponding hold-out test sets, and shaded areas correspond to two standard deviations from the respective means.}
\label{fig:perf_curves_length_of_stay}
\end{figure}

In addition to reducing the number of unsuccessful discharges, $\mu_{OP}$ can  identify patients that can be safely discharged at a much earlier stage in their ICU stay than $\mu_{CP}$. For low values of $g(UD)$, $\mu_{OP}$ is expected to favor discharges that minimize hospitalization costs, thus reducing the patients' length of stay. This property of $\mu_{OP}$ is demonstrated in Figure \ref{fig:perf_curves_length_of_stay}, showing the resulting fraction of $UD$ and the average length of stay for same range of values of  $g(UD)$ used in Figure \ref{fig:perf_curves_OP_RP_CP}. For instances in which $\mu_{OP}$ recommended delayed discharge relative to $\mu_{CP}$, 100 transitions were simulated from the last health state observation to compute the expected length of stay (which is naturally higher than that of $\mu_{CP}$). The corresponding results for $\mu_{RP_2}$ relative to $\mu_{CP}$ are not shown since the timing of discharge of $\mu_{CP}$ was the same as that of $\mu_{CP}$, and thus $\mu_{RP_2}$ had an average expected length of stay greater than or equal to the one observed for $\mu_{CP}$. Interestingly, Figure \ref{fig:perf_curves_length_of_stay} suggests that $\mu_{OP}$ obtained using $g(UD) \sim 2$ reduced the mean expected length of stay by 0.88 days ($p$-value $< 0.001$) while also reducing the mean percentage of unsuccessful discharges by 2.00\% ($p$-value $< 0.001$), relative to the mean values observed for $\mu_{CP}$. Letting $g(UD) \sim 3$ $\mu_{OP}$ (corresponding to the cost function chosen for the results in Section \ref{ss:cost_eval}) yielded an increase of the mean patient length of stay of 2.4 days ($p$-value $< 0.001$) and a reduction of the mean percentage of unsuccessful discharges of 2.83\% ($p$-value $< 0.001$). 

From Figure \ref{fig:perf_curves_length_of_stay} it is clear that the two objectives (i.e., the costs associated with hospitalization and unsuccessful discharges) are conflicting, since reducing the number of patients that are readmitted or deceased after discharge can only be accomplished by increasing their length of stay (an implied medical treatment) in the ICU. The parameter $g(UD)$ can be used by hospital management and clinicians to address their preferences associated with the two cost factors, reflecting aspects such as hospital capacity and resource availability at a given period in time. Despite the conflicting objectives, these results show that $\mu_{OP}$ improved upon the clinician policy with respect to the two metrics in the hold-out test sets.

\subsection{Policy interpretability}
\label{ss:interpretability}

\begin{figure}[htb]
\centering
\includegraphics[scale=0.45]{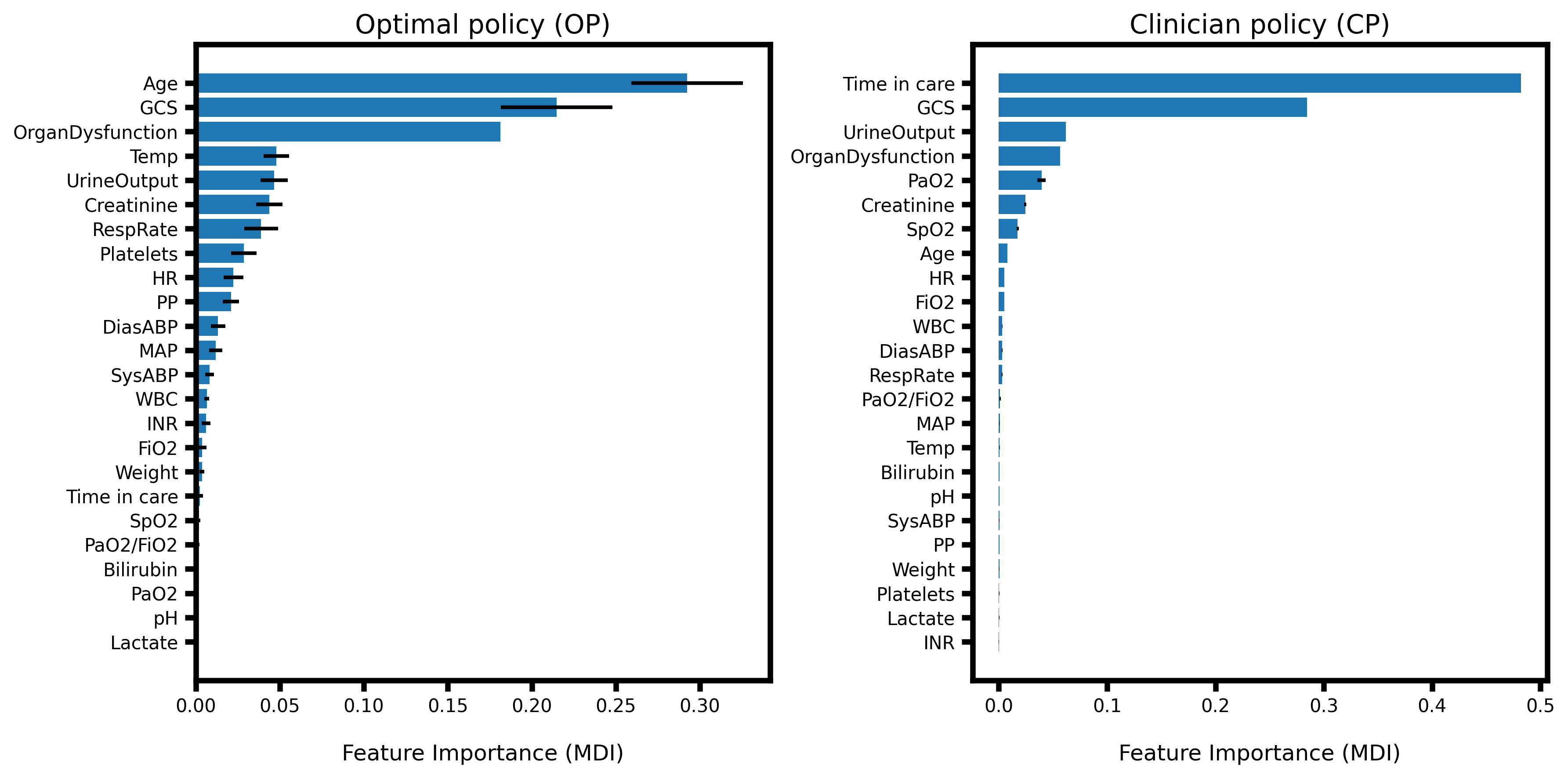}
\caption{Average feature importances corresponding to classifiers trained to predict $\mu_{OP}$ and $\mu_{CP}$, for all 50 train-test splits with associated  clustering models and optimal policies identified using $g(UD)=3$. Error bars correspond to the feature importance standard deviation over the 50 train-test splits.}
\label{fig:feature_importance}
\end{figure}

To better understand the difference in the factors that drive the clinician and proposed policies, classifiers were trained to predict discharge decisions associated with each policy based on the full-dimensional data (i.e., using the complete patient's EHR without any clustering or dimensionality reduction). The binary classification task at hand used the full set of clinical and demographic features observed at a point in time to predict whether a patient was discharged or not by a given policy. A gradient-boosted decision trees classifier (maximum tree depth of 4, and number of estimators of 100) was used to predict the discharge outcome for each policy on the same training set used to perform clustering to generate the patient's health state. The receiver operating characteristic (ROC) curve for the classifiers and corresponding area under the curve (AUC) for the two policies are shown in Figure S6, indicating that both policies can be predicted with high performance (median AUC of 0.937 and 0.920 for $mu_{OP}$ and $\mu_{CP}$ respectively) using the full-dimensional data. The feature importances corresponding to mean decrease in impurity (MDI) or Gini importance for each of the classifiers are shown in Figure \ref{fig:feature_importance}, which serve to shed light on to main factors being considered by each policy.

The average feature importances shown in Figure \ref{fig:feature_importance} confirm that the discharge decisions corresponding to $\mu_{OP}$ are clinically interpretable and rely on sensible clinical features. In particular, several of the most important features used in predicting the clinician policy also have high importance in predicting the proposed optimal policy, which further supports its interpretability. Nonetheless, the feature importances corresponding to $\mu_{OP}$ reveal that the proposed policy employs a much larger set of relevant clinical features in recommending patient discharge than $\mu_{CP}$. This property enables $\mu_{OP}$ to discern patients that are at high risk of being unsuccessfully discharged, thus leading to an overall improved policy relative to the clinician policy and the random policies used as benchmarks.  

\section{Limitations and future work}

An inherent limitation of the proposed policy learning framework is that the MDP's Markov property may prevent the leveraging of a patient's past clinical history when making the discharge decision. In particular, we note that $\mu_{OP}$ may result in discharging patients that show a brief, temporary improvement in their health condition but later exhibit a rapid deterioration and in-hospital death. These results are shown in Figure S5, by evaluating $\mu_{OP}$ on a hold-out test set with in-hospital death (IHD) instances. While, at the moment of discharge, future IHD and non-IHD instances are almost indistinguishable based on the respective mortality risks, it is evident that: (i) patients with IHD were observed to be in clusters with higher mortality rates than non-IHD patients prior to $\mu_{OP}$ recommending discharge, and (ii) the transitions following $\mu_{OP}$ discharge recommendation also correspond to clusters with much higher mortality rates. These observations suggest that $\mu_{OP}$ can be employed in practical settings in combination with the clinician's assessment of the patients clinical history, and with a ``wait-and-see'' period in which the patient is kept in care for a brief observation period. From a modeling perspective, it would be interesting to explore higher-order Markov models which consider explicit dependencies on prior system states. Additional features reflecting prior health measurements could also be incorporated in the state clustering procedure to implicitly include a patient's clinical history.

Moreover, based on the structural elements introduced in Section \ref{ss:MDP_structure} the resulting MDP is time-homogeneous with respect to the transition probabilities and the costs per stage (i.e., these components do not vary in time). Certain practical settings where the transition probabilities might be time-varying depending on a patients length of stay would require formulating time-inhomogeneous MDPs. These models have been much less studied in the literature relative to their time-homogeneous counterparts, and their application to clinical decision-making problems is an interesting direction of future research.

We note that despite the proposed framework's superior performance relative to clinician discharge decisions, the resulting policy is still a proof-of-concept and does not constitute a medical device that is ready for deployment. The proposed framework should be validated prospectively to a greater extent on observational data sets collected from multiple different hospitals and in a variety of clinical settings. For clinical use, extensive clinical trials must be performed to evaluate this work in real-time, in closed-loop, and interactively with ICU patients \cite{komorowski2018artificial}. Furthermore, it is clear that the implementation of clinical decision support tools is hindered by several factors including their compatibility and interoperability with current EHR systems, inconsistency and hospital-specific
practices in recording and storing of patient data, and concerns of eventual over-reliance on models over
medical expertise \cite{sutton2020overview}. Based on these limitations it is evident that on top of the additional validation studies required, the proposed tool must always be used in conjunction with physicians' subjective judgement about patient discharge decisions and treatment strategies.

\section{Conclusions}

The development of clinical decision support tools can play a significant role in improving the outcome of ICU patient discharges, as well as in improving the overall hospital management efficiency. In this paper, we presented an end-to-end framework for modeling patient health states to subsequently derive data-driven policies for optimal patient discharge recommendation. The proposed policies balance competing objectives relating to hospitalization costs and penalties associated with unsuccessful discharges (i.e., $T$-day ICU readmission or out-of-hospital mortality). Extensive numerical experiments performed on real-life ICU patient EHR data demonstrated that the proposed policy reduced both the rate of readmission and the average patient length of stay for certain cost parameters. Furthermore, the off-policy evaluation strategies showed that the proposed policy consistently outperformed the clinician and random policy benchmarks. At the same time, the policy was shown to be clinically interpretable, an auspicious property for its deployment in practical settings. 


\section*{Acknowledgements}

We acknowledge the insightful comments provided by Dr. Jana Hoffman in writing the manuscript. 

\section*{Competing interests}
All authors who have affiliations listed with Dascena (Houston, Texas, U.S.A) are employees or contractors of Dascena




\begin{singlespacing}
\bibliographystyle{elsarticle-harv} 
\bibliography{discharge_paper_arxiv}
\end{singlespacing}

\end{document}


\renewcommand{\theequation}{S\arabic{equation}}
\renewcommand{\thefigure}{S\arabic{figure}}
\renewcommand{\thetable}{S\arabic{table}}
\renewcommand{\thealgocf}{S\arabic{algocf}}

\title{\textbf{Supporting Information}: Optimal discharge of patients from intensive care via a data-driven policy learning framework}

\author[1,2]{Fernando Lejarza} 
\ead{flejarza@dascena.com} 
\author[1]{Jacob Calvert}
\ead{jake@dascena.com} 
\author[1]{Misty M Attwood} 
\ead{mattwood@dascena.com} 
\author[1]{Daniel Evans} 
\ead{devans@dascena.com} 
\author[1]{Qingqing Mao} 
\ead{qingqing@dascena.com} 
\address[1]{Dascena, Inc., 12333 Sowden Road, Suite B, Houston, TX 77080-2059, USA\fnref{label1}}
\address[2]{McKetta Department of Chemical Engineering, The University of Texas at Austin, Austin, TX 78712, USA\fnref{label2}}
\maketitle

\section{Data processing}

\begin{figure}[htb]
\centering\includegraphics[scale=0.5]{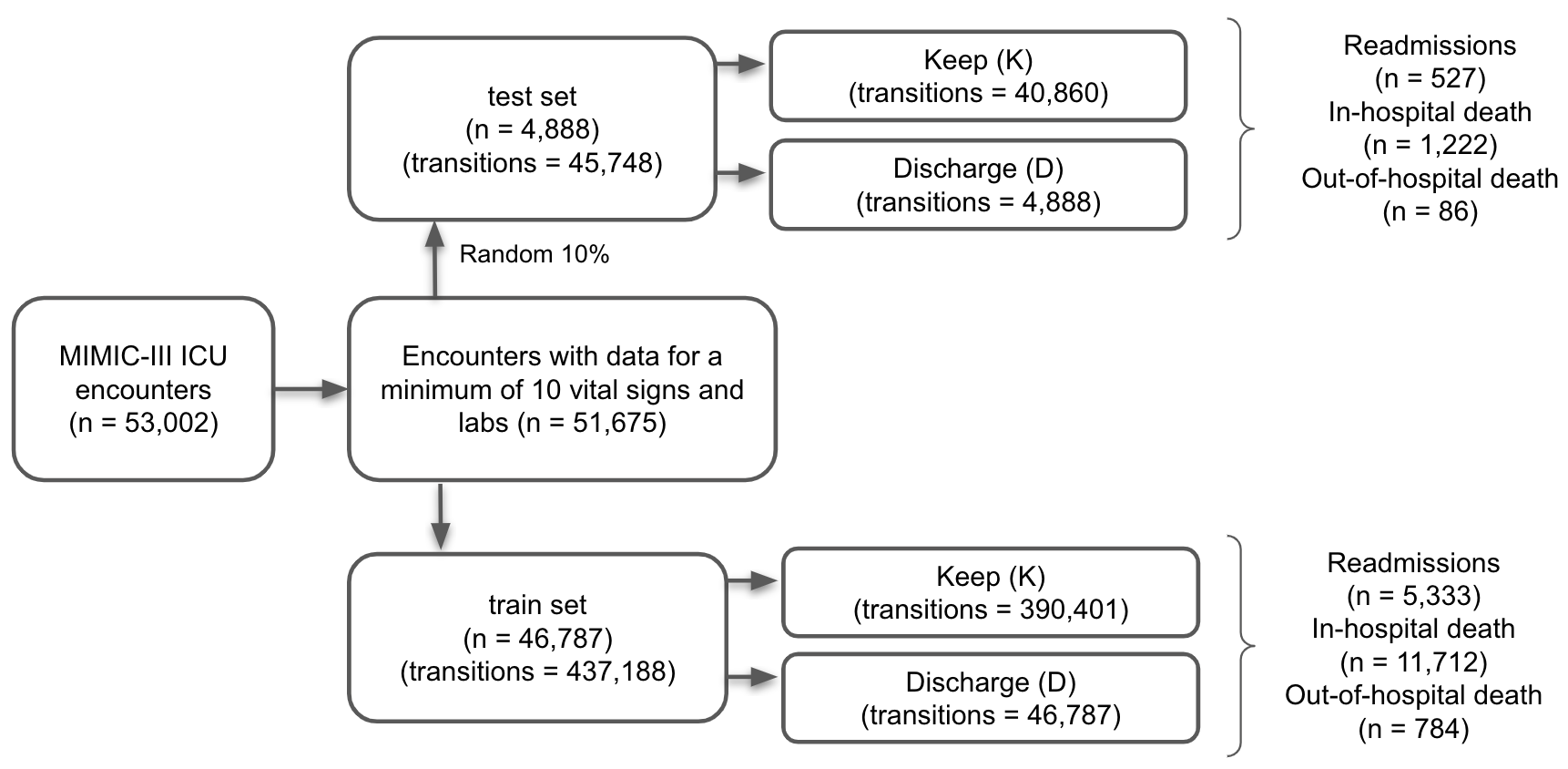}
\caption{Patient inclusion diagram ($n$ corresponds to the number of ICU stays, and transitions corresponds to the total number of 12-hour window observations for all ICU stays).}
\label{fig:inclusion}
\end{figure}

\begin{table}[H]
\caption{Demographics of the included MIMIC-III ICU stays. The ICU types are mutually exclusive and include: medical intensive care unit (MICU), surgical intensive care unit (SICU), coronary care unit (CCU), cardiac surgery recovery unit (CSRU), trauma-surgical intensive care unit (TSICU). IQR denotes the interqaurtile range.}
\centering
\label{tb:features} 
\begin{tabular}{p{0.35\linewidth}  p{0.15\linewidth}  p{0.45\linewidth}}
\hline
\multicolumn{2}{c}{\textbf{Demographic characteristic}} & \textbf{Number of ICU stays $n$ (\%)} \\
\hline 
\textit{ICU type} & MICU  & 20172 (39.036) \\
 &                 SICU & 8669 (16.78) \\
 &                 CCU  & 7513 (14.54) \\
 &                 CSRU &  9593 (18.56)  \\
 &                 TSICU & 5728 (11.08) \\
 & \\ 
\textit{Gender} & Female & 22561 (43.65)\\ 
 &                 Male & 29114 (56.34) \\ 
 & \\
\textit{Age (years)} & 15-19 & 385 (0.75)  \\ 
 Median 66.00 & 20-29 & 2044 (3.96) \\ 
IQR (53.00-77.00)& 30-39 & 2711 (5.25) \\ 
& 40-49 & 5485 (10.61) \\ 
& 50-59 & 9043 (17.50) \\ 
& 60-69 & 10873 (21.04) \\ 
& 70-79 & 10663 (20.63) \\ 
& 80-90 & 10471 (20.26) \\ 
& \\ 
\textit{Length of stay (days)} &  0-2 & 23303 (45.10) \\
 Median 7.00 &                 3-5  & 17498 (33.86) \\ 
 IQR (3.00-17.00) &                 6-8 & 4623 (8.95) \\
 & 9-11 &  2138 (4.14) \\ 
 & 12+ &  4113 (7.96) \\ 
& \\ 
 \textit{Readmission$^*$} & Yes & 5860 (11.34) \\ 
 &                 No  &  45815 (88.66)\\
 \textit{In-hospital death} & Yes  &  15445 (29.89)  \\ 
 &                 No & 36230 (70.11) \\ 
 & \\ 
  \textit{Out-of hospital death$^*$} & Yes & 870 (1.68)  \\ 
 &                 No  &  50805 (98.32) \\
\hline 
\multicolumn{3}{l}{$^*$Readmitted or dead within 30 days from discharge}
\end{tabular} 
\end{table}

\begin{table}[H]
\caption{Data included as input for learning discrete patient health state representation (BP: blood pressure, GCS: Glasgow Coma Scale, HR: heart rate, SpO\textsubscript{2}: oxygen saturation, PP: pulse pressure , MAP: mean arterial pressure, FiO\textsubscript{2}: fraction of inspired oxygen, INR: international normalized ratio, WBC: white blood cell count, PaO\textsubscript{2}: partial pressure arterial oxygen) }
\centering
\label{tb:features} 
\begin{tabular}{p{0.25\linewidth}   p{0.74\linewidth}}
\hline
\textbf{Category} & \textbf{Features} \\ 
\hline
\multirow{3}{*}{\textit{Vital signs and labs}} & Diastolic BP, Systolic BP, GCS, HR, Respiration rate, SpO\textsubscript{2}, Temperature, PP, MAP, Bilirubin, Creatinine, FiO\textsubscript{2}, INR, Lactate, pH, Platelets, WBC, PaO\textsubscript{2}, Urine output, PaO\textsubscript{2}/FiO\textsubscript{2}, \\   
& \\ 
\multirow{2}{*}{\textit{Treatments}} & Antibiotics, Fluid bolus, Norepinephrine, Dopamine, Vasopressin, Dobutamine, Phenylephrine, Epinephrine \\  
 & \\ 
\textit{Demographics }& Age, Weight \\
& \\ 
\textit{Other} & Time period in care, Organ dysfunction (boolean) \\ 
\hline 
\end{tabular} 
\end{table}

\section{Additional results for patient health state clustering}

\begin{figure}[htb]
\centering\includegraphics[scale=0.45]{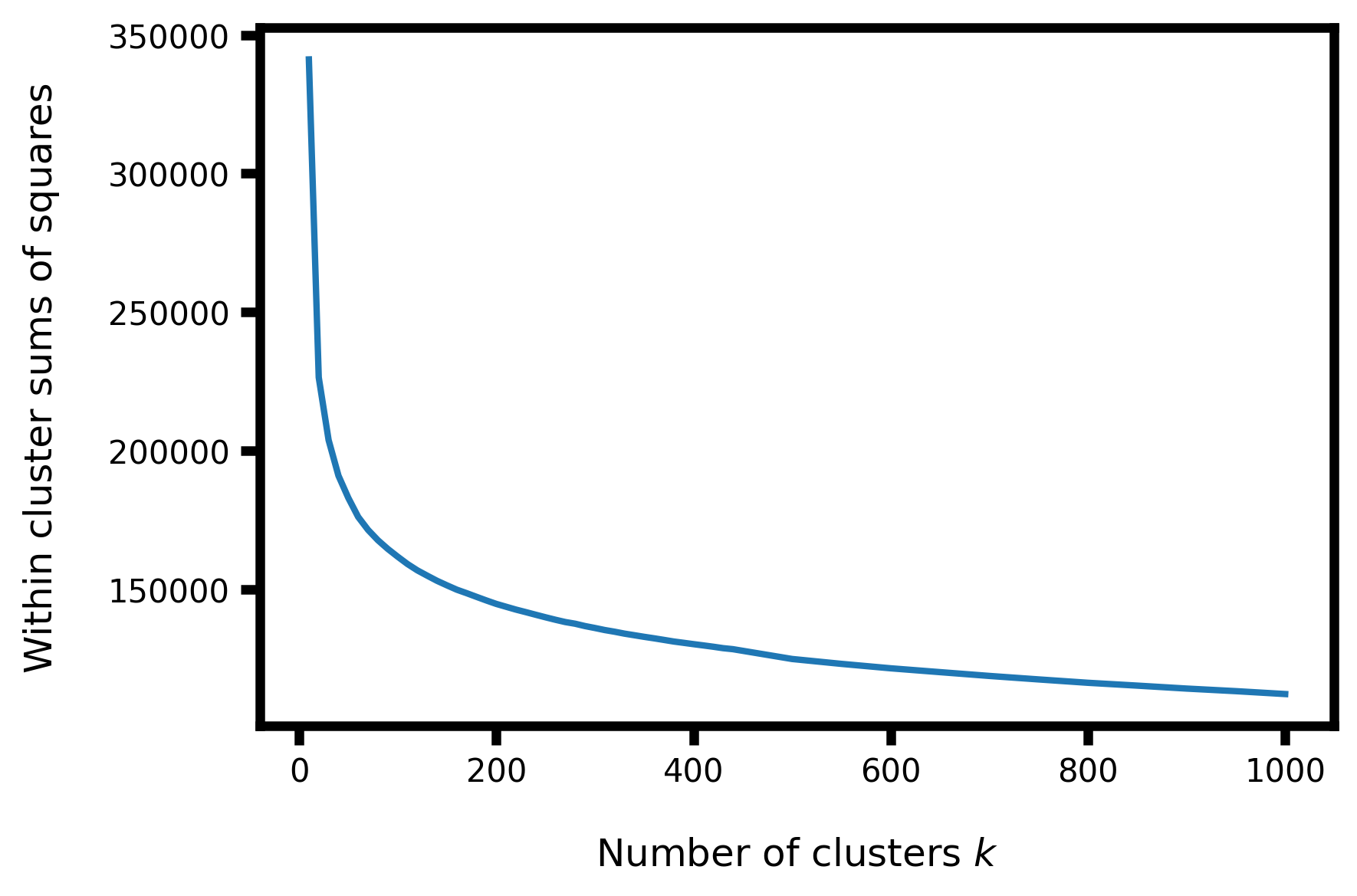}
\caption{Elbow method showing the within cluster sums of squares as a function of the number of clusters in $k$-means algorithm.}
\label{fig:elbow_method}
\end{figure}

\begin{table}[htb]
\caption{$R^2$ values for 50 different clustering instances and associated transition probability estimates. Values shown correspond to the median, and the interquartile range (IQR) is shown in parenthesis.}
\centering
\label{tb:opt_instances} 
\begin{tabular}{c c}
\hline
\textbf{Transition probability} &\textbf{ \textbf{$R^2$}}  \\  
\hline 
$P(j|i,K)$ & Median: $0.947$, IQR: $(0.945 - 0.949)$ \\  
$P(SD|i,D)$ & Median: $0.974$, IQR: $(0.971 - 0.976)$ \\  
$P(UD|i,D)$ & Median: $0.845$, IQR: $(0.831 - 0.859)$ \\
\hline 
\end{tabular} 
\end{table}

\begin{figure}[htb]
\centering
\subfloat[]{\label{fig:A}\includegraphics[width=0.5\linewidth]{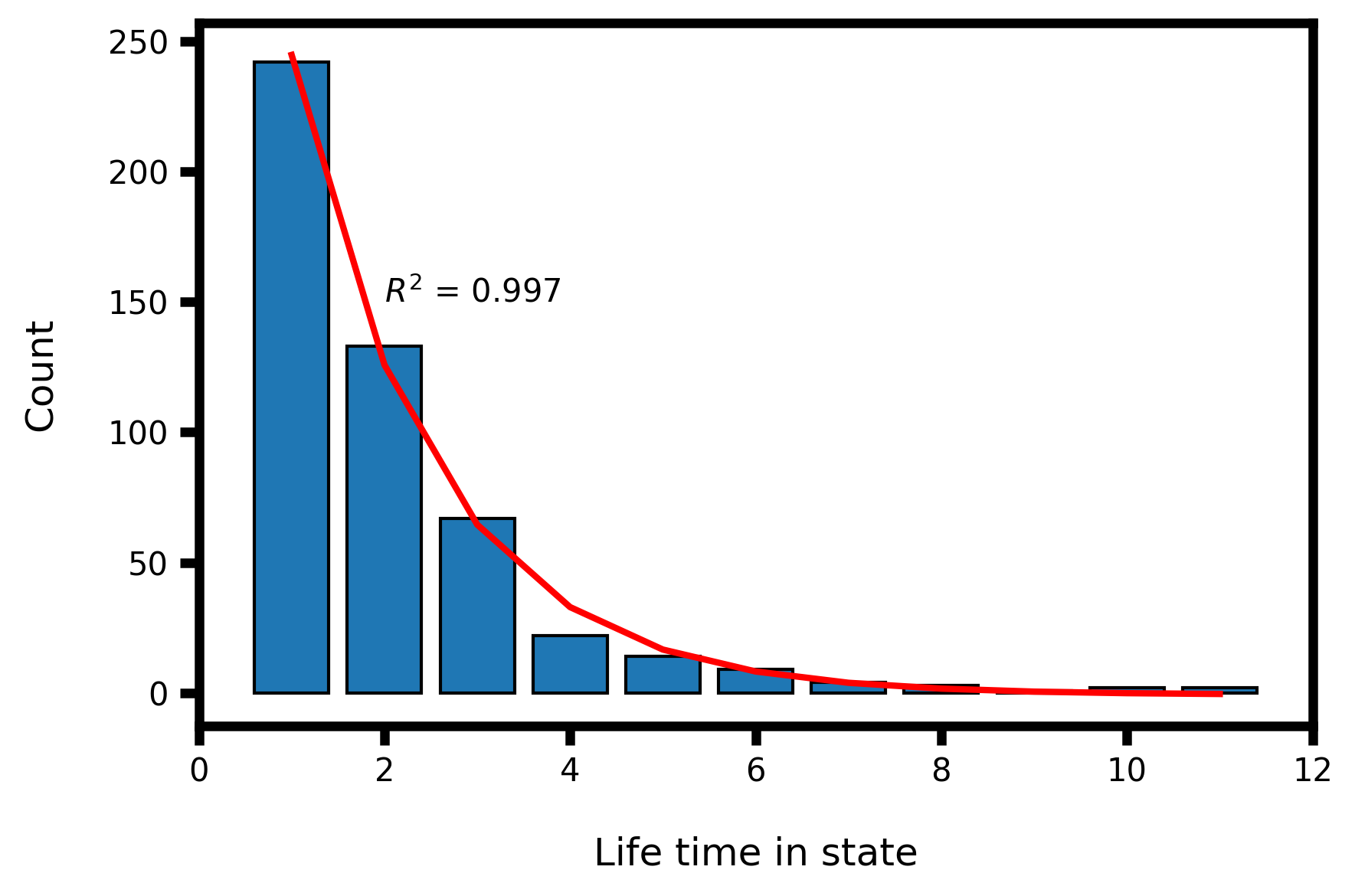}}\hfill
\subfloat[]{\label{B}\includegraphics[width=.5\linewidth]{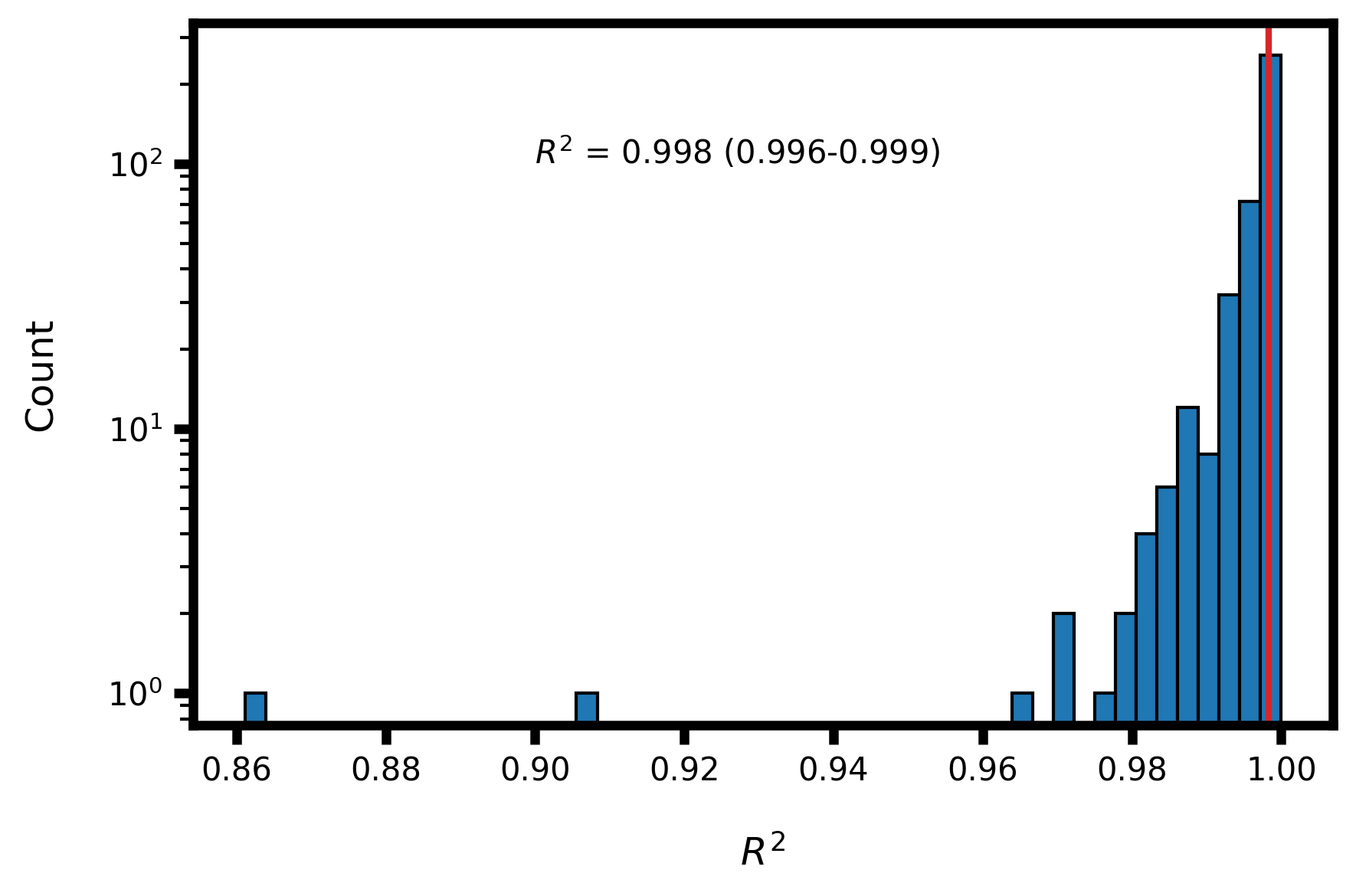}}\par 
\caption{Verifying memoryless Markovian property for the estimated transition probabilities using 500 simulations starting from every initial state. (a) Lifetime distribution for a given state and fitted exponential decay function of the form of $f(x)=\gamma e^{-\lambda x}$. (b) Distribution of $R^2$ coefficients between fitted models and observed life times for all 400 clusters, text indicates the median $R^2$ and interquartile range in parenthesis.}
\label{fig:markov_property}
\end{figure}

\section{Additional numerical results for policy assessment}

\begin{figure}[htb]
\centering
\includegraphics[scale=0.45]{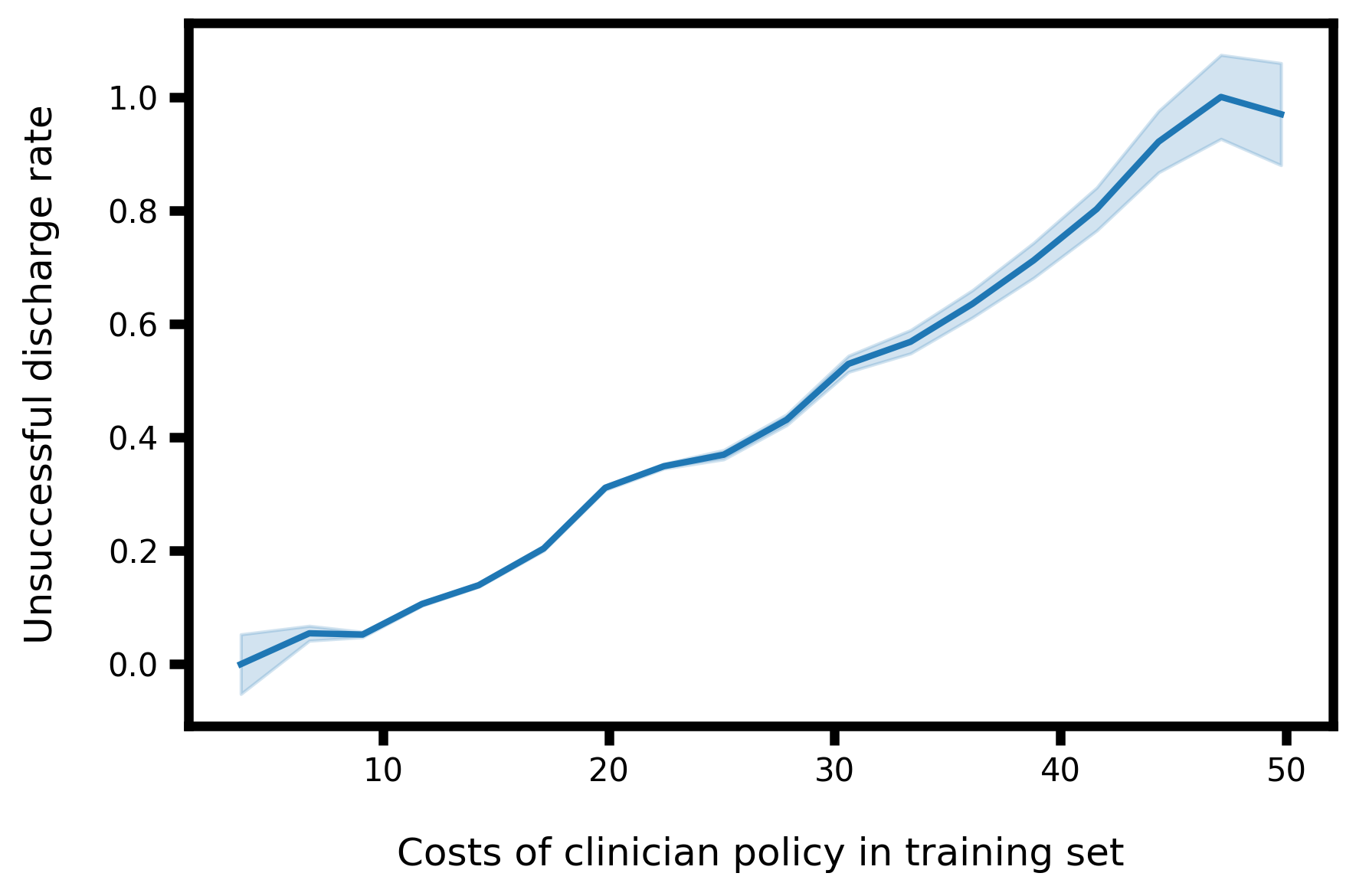}
\caption{Policy calibration showing the rate of unsuccessful discharge related to the costs incurred by the clinician policy using $g(UD)$.}
\label{fig:policy_calibration}
\end{figure}

\begin{table}[htb]
\caption{Off-policy evaluation results corresponding to different number of clusters used to derive $\mu_{OP}$.  Values shown correspond to the mean confidence bound of the policy cost for five different models trained for each number of clusters, and the standard deviation is shown in parenthesis.}
\centering
\label{tb:cluster_number} 
\begin{tabular}{c c c}
\hline
\textbf{Number of clusters} & $\mu_{OP}$ 95\% UB   &  $\mu_{CP}$ 95\% LB  \\  
\hline 
350 & 16.81 (0.52) & 18.41 (0.16) \\  
400 & 16.17 (0.65) & 18.36 (0.16)  \\  
450 & 15.50 (0.63) & 18.20 (0.17)  \\
\hline 
\end{tabular} 
\end{table}

\begin{figure}[htb]
\centering
\subfloat[]{\label{fig:A}\includegraphics[width=0.5\linewidth]{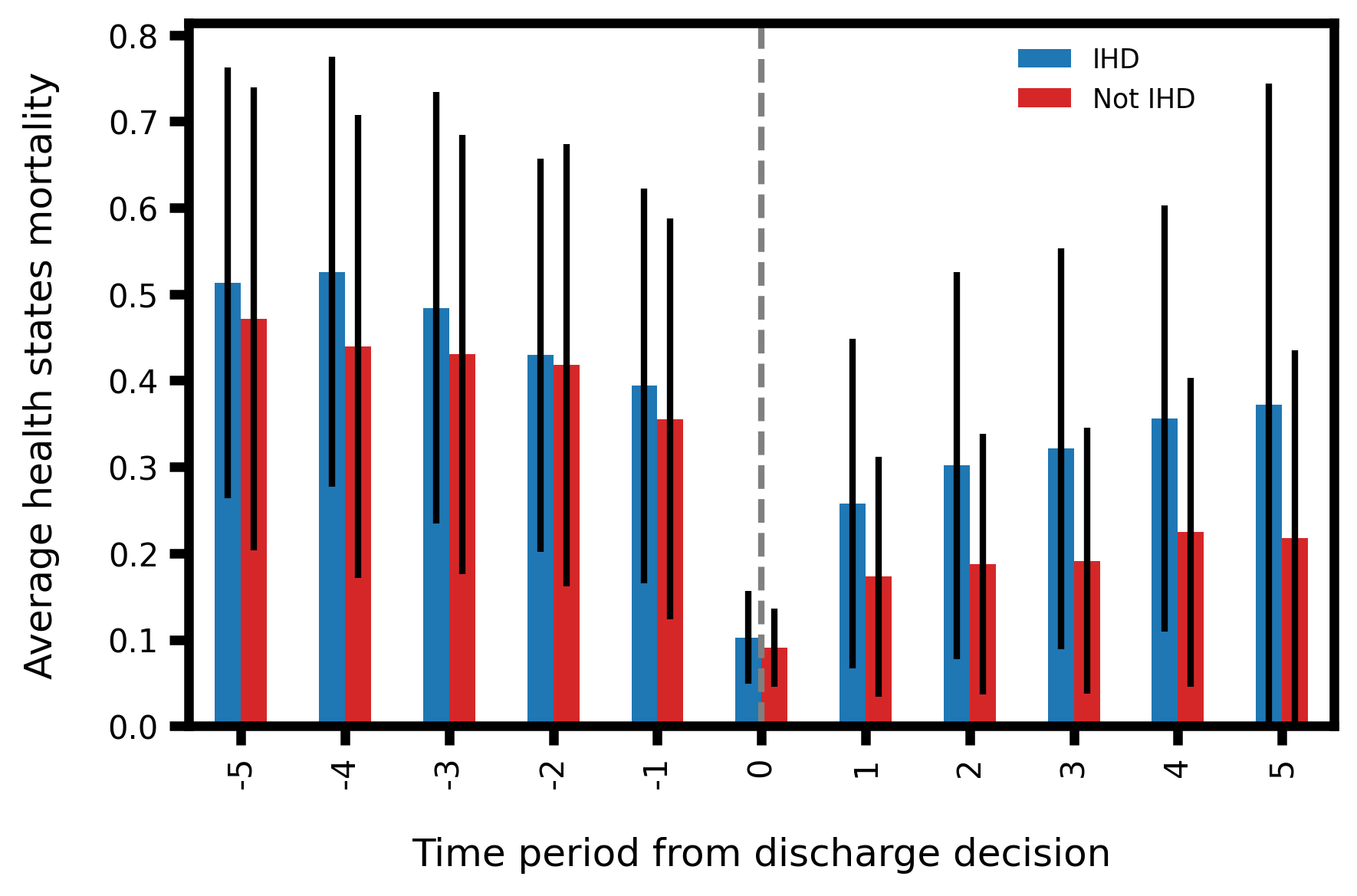}}\hfill
\subfloat[]{\label{B}\includegraphics[width=.5\linewidth]{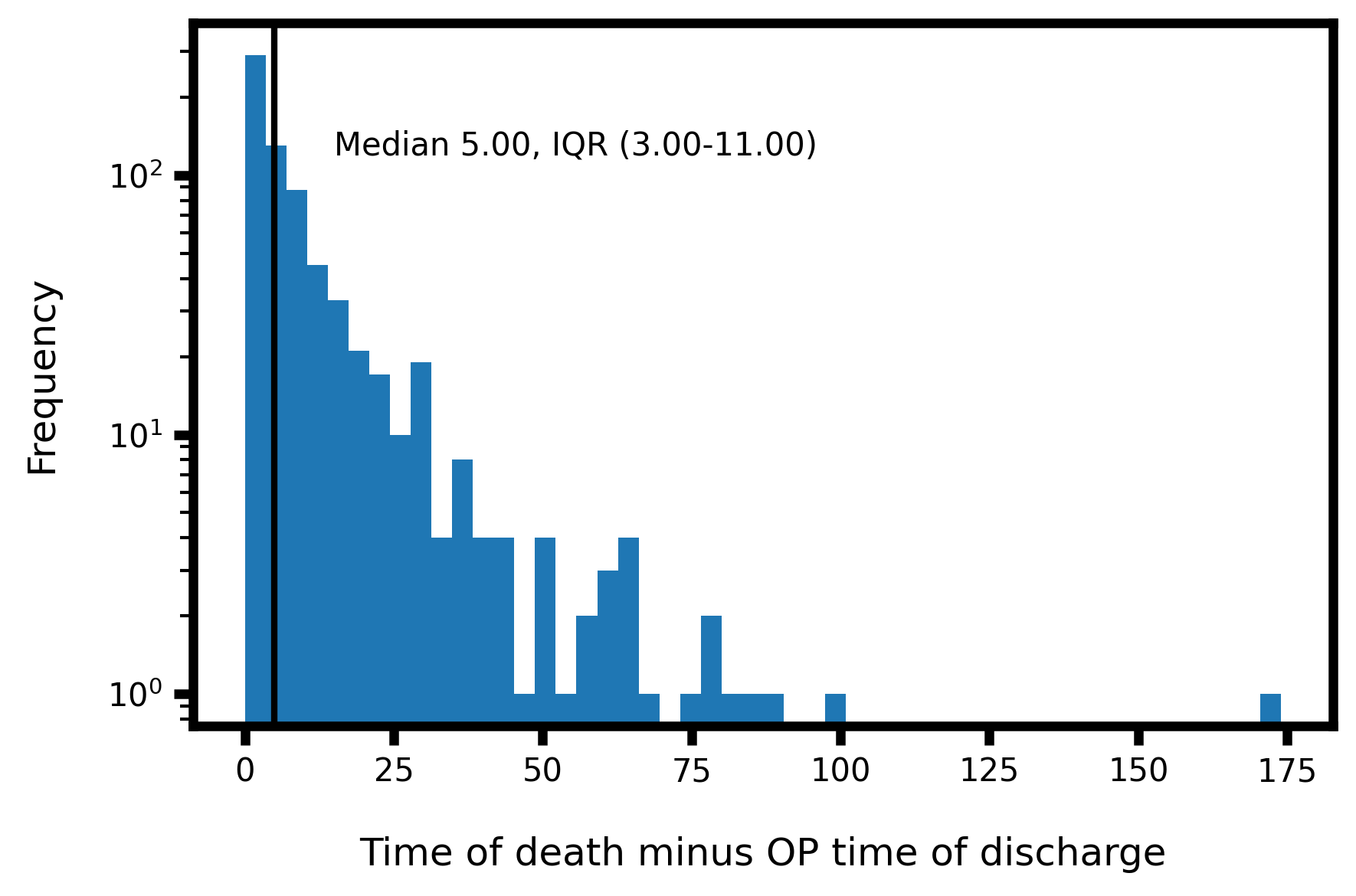}}\par 
\caption{(a) Average observed cluster mortality as a function of time relative to the moment when OP recommends discharge at time $t=0$, for instances in a single hold-out test set (error bars correspond to the standard deviation). OP was generated using cost parameter $g(UD)=3$., (b) Distribution of time difference between observed time of death and timing of discharge by OP in time period (vertical black line corresponds to the sample median).}
\label{fig:elbow_method}
\end{figure}

\begin{figure}[htb]
\centering
\includegraphics[scale=0.45]{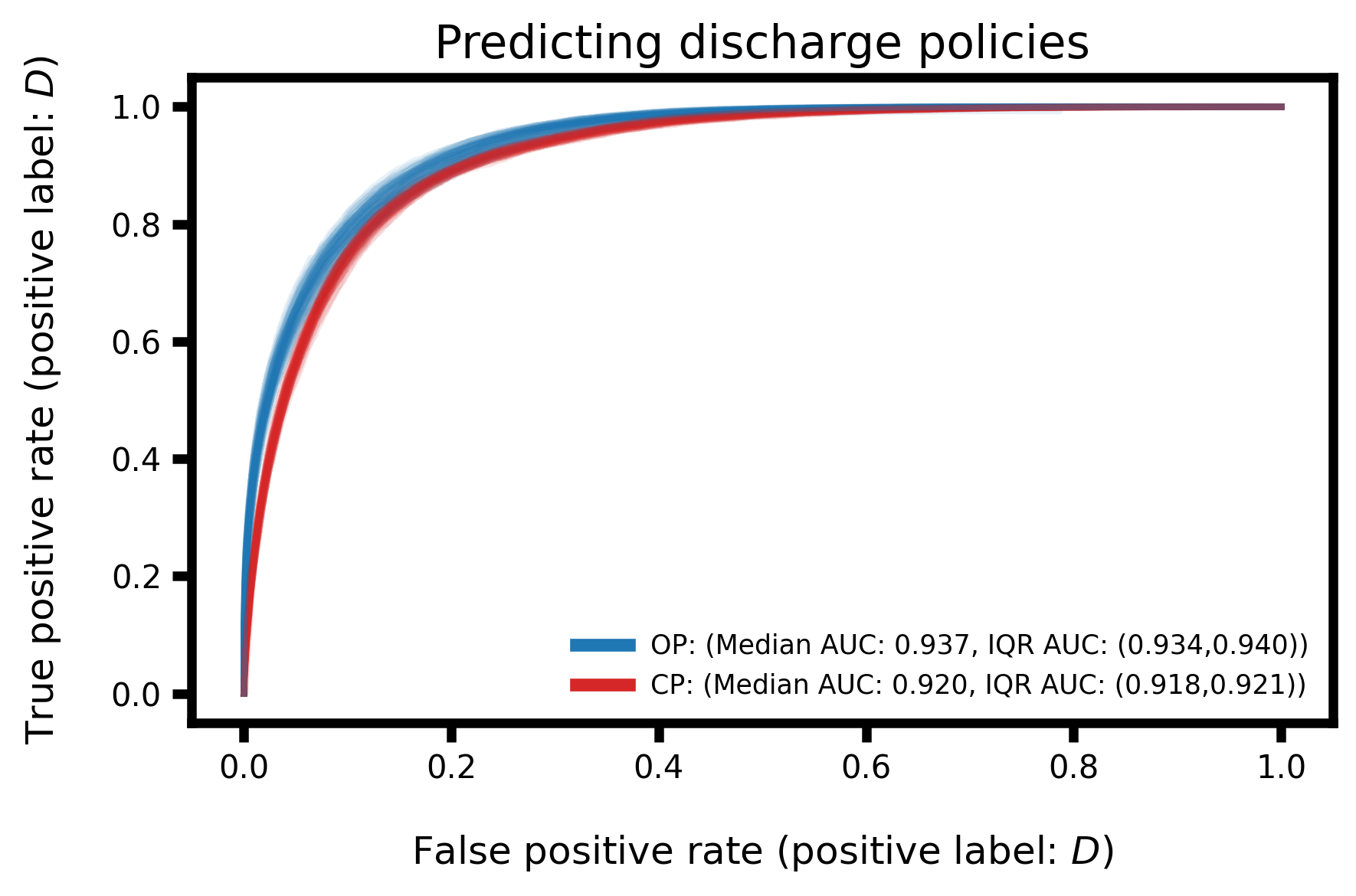}
\caption{Receiver operating characteristic (ROC) curve for all  50 models developed to predict $\mu_{OP}$ and $\mu_{CP}$.}
\label{fig:roc}
\end{figure}